\documentclass[journal]{IEEEtran}
\usepackage{amsmath,amssymb,bm}
\usepackage{array,threeparttable,booktabs,multirow}
\usepackage{graphicx}
\usepackage[ruled]{algorithm2e}
\usepackage[noend]{algpseudocode}
\usepackage{cite}
\usepackage[dvipsnames]{xcolor}
\usepackage{setspace}
\usepackage[hidelinks]{hyperref}
\usepackage{pifont}
\usepackage{makecell}
\usepackage{diagbox}
\usepackage{setspace}
\usepackage{float}   
\usepackage{subfigure}

\begin{document} 
\bstctlcite{BSTcontrol}
\linespread{0.85}

\title{
	Detection and Imputation based Two-Stage Denoising Diffusion Power System Measurement Recovery under Cyber-Physical Uncertainties
}
\author{
	Jianhua~Pei,~\IEEEmembership{Student~Member,~IEEE}, %
	Jingyu~Wang,~\IEEEmembership{Member,~IEEE}, %
	Dongyuan~Shi,~\IEEEmembership{Senior~Member,~IEEE}, %
	and~Ping~Wang,~\IEEEmembership{Fellow,~IEEE} %

	\thanks{
		Manuscript received December 7, 2023; revised April 5, 2024; accepted May 27, 2024. This work was supported by the National Natural Science Foundation of China (Grant No. 52207107) and China Scholarship Council (Grant No. 202306160010). Paper no. TSG-02069-2023. 
		\emph{(Corresponding author: Jingyu Wang)}
	}%
	\thanks{
		J. Pei, J. Wang and D. Shi are with the State Key Laboratory of Advanced Electromagnetic Technology, School of Electrical and Electronic Engineering, Huazhong University of Science and Technology, Wuhan, China (e-mails: \{\href{jianhuapei@hust.edu.cn}{jianhuapei}, \href{jywang@hust.edu.cn}{jywang}, \href{dongyuanshi@hust.edu.cn}{dongyuanshi}\}@hust.edu.cn). P. Wang is with the Department of Electrical Engineering and Computer Science, Lassonde School of Engineering, York University, Toronto, Canada (e-mail:pingw@yorku.ca).
	}%

}
\maketitle

\begin{abstract}
	Power system cyber-physical uncertainties, including measurement ambiguities stemming from cyber attacks and data losses, along with system uncertainties introduced by massive renewables and complex dynamics, reduce the likelihood of enhancing the quality of measurements. Fortunately, denoising diffusion models exhibit powerful learning and generation abilities for the complex underlying physics of the real world. To this end, this paper proposes an improved detection and imputation based two-stage denoising diffusion model (TSDM) to identify and reconstruct the measurements with various cyber-physical uncertainties. The first stage of the model comprises a classifier-guided conditional anomaly detection component, while the second stage involves diffusion-based measurement imputation component. Moreover, the proposed TSDM adopts optimal variance to accelerate the diffusion generation process with subsequence sampling. Extensive numerical case studies demonstrate that the proposed TSDM can accurately recover power system measurements despite renewables-induced strong randomness and highly nonlinear dynamics. Additionally, the proposed TSDM has stronger robustness compared to existing reconstruction networks and exhibits lower computational complexity than general denoising diffusion models.
\end{abstract}


\section{Introduction} % (fold)
\label{sec:introduction}

\IEEEPARstart{M}{odern} power systems typically exhibit a high level of complex nonlinear dynamics, high volatility, randomness, and intermittency due to the integration of renewable energy and the existence of interconnected cyber-physical power system events \cite{10459229}. In light of this, the new electric system has raised more concerns about the data quality issues of deployed remote terminal units (RTUs) in supervisory control and data acquisition (SCADA) and phasor measurement units (PMUs) in wide-area measurement systems (WAMS), as the accurate power system state variables estimated by RTU and PMU measurements often serve as input for many energy management applications \cite{7442164}. However, measurement uncertainties, such as communication contingencies in extreme weather events and cyber-attacks in malicious threats, severely degrade the quality of power system measurements \cite{9815319}. Specifically, accidental network congestion or deliberate communication blocking can lead to data losses in the control center and manufactured data tampering attacks can cause the estimated system states to be biased, leading to the control center's incorrect decisions \cite{7873357}. For this reason, adequate attention should be paid to investigating data recovery approaches against corrupted fragments in SCADA and WAMS.

In SCADA, RTUs collect the measurements of power flow, power injection, and voltage magnitude at a low sampling frequency and then upload them to the control center. Nowadays, most power utilities still use weighted least squares-based static state estimation (SSE) to estimate the system's state variables, i.e., the bus voltage phasors. The residuals are calculated to identify and eliminate bad data \cite{4803729}. Moreover, the measurement redundancy \cite{kotha2022power} and Lagrange interpolation method \cite{huang2016data} can handle a small amount of data losses. Nonetheless, the emergence of data manipulation attack (DMA) represented by false data injection attack (FDIA) has demonstrated the cyber security vulnerability of the above methods, i.e., FDIA can covertly inject false data vectors and seriously degrade the power system data quality \cite{10013729}. Most of the subsequent studies on DMA focus on how to launch attacks stealthily and on defense mechanisms, and only a minority of the literature considers data recovery after suffering from manipulation. The generative adversarial network (GAN) model is employed to generate non-tampered data for recovering the compromised state variables \cite{8731755}. A P-Q decomposition-based linear transformation approach is used to reconstruct contaminated state variables \cite{9117029}. Nonetheless, the nonlinear nature of power flow may reduce the recovery accuracy. In \cite{9055170}, a state recovery approach is proposed based on the combination of quasi-Newton and Armijo line searches. However, successfully identifying the contaminated data by additional extreme learning machines is still necessary. The inertia of the power system is utilized for data restoration against FDIA with high accuracy and efficiency \cite{9697341}. As the renewables-penetrated power systems may be low-inertia, the recovery error is high when state variables fluctuate dramatically.

The PMUs are deployed in WAMSs for real-time monitoring power systems at high or ultra-high sampling frequencies. Dynamic state estimation (DSE) methods, represented by extended Kalman filtering (EKF) \cite{ghahremani2011dynamic}, unscented Kalman filtering (UKF) \cite{qi2016dynamic}, and particle filtering (PF) \cite{abolmasoumi2023robust}, are employed to track the internal states of generators and filter the PMU measurements using the correlation between adjacent sampling instants. However, these commonly utility-deployed DSE methods exhibit low estimation accuracy when encountering gross outliers induced by FDIA and PMU data manipulation attack (PDMA) \cite{wang2019online} or PMU signal losses. To this end, numerous improved Kalman filtering-based methods have been proposed to enhance robustness against anomalies and data losses \cite{zhao2016robust, taha2016risk, zhao2019correlation}. However, the estimation results of these methods may still deviate from normal states when facing well-designed malicious injected false data. Additionally, methods that utilize the low-rank property of PMU time series are developed to limitedly recover the missing data \cite{7080935,pourramezan2020real,9459703}. Among them, the alternating direction method of multiplier (ADMM) \cite{8428530} is utilized to improve the recovery accuracy and convergence speed only when the measurements are accurately observed. Machine learning-based approaches using gated recurrent units (GRU) \cite{8356714} and graph neural networks (GNN) \cite{8642404} are another type of method that can get higher recovery accuracy. However, this class of learning methods, especially time series prediction-based models represented by long short-term memory networks (LSTM) and GRU, may have overfitting and generalization issues when confronting strong randomness and complex nonlinearity in new-generation electric systems. Furthermore, mathmatical algorithms based on robust principal component analysis (RPCA) \cite{8120129}, subspace selection \cite{chatterjee2019robust}, k-medoids clustering \cite{9946434}, singular value decomposition \cite{pei2022robust}, and Tucker decomposition \cite{osipov2020pmu} can effectively address PMU data anomalies or missing data issues, contributing to the cyber-resilient power systems. Unfortunately, all these existing recovery approaches may exhibit accuracy bottlenecks in the presence of cyber-physical uncertainties.

The existing power system data recovery methods for DMA and data loss issues still need improvement, such as improving recovery accuracy, expanding application scenarios, and reducing computational complexity. Compared with existing methods, denoising diffusion models, represented by the denoising diffusion probabilistic model (DDPM) \cite{2020DDPM} and the denoising diffusion implicit model (DDIM) \cite{2020Denoising}, are better at generating realistic measurements with strong randomness and support potential guided conditional data generation, making them more suitable for data recovery than existing generative models in practical power systems. However, the controllable data generation and efficient computing methods of diffusion models for deterministic data recovery trajectory still need further investigation. In this paper, a enhanced two-stage denoising diffusion recovery model (TSDM) is proposed to recover power system measurements under DMAs and data losses. The overall contribution of this approach is threefold:

\begin{enumerate}
	\item A novel detection and imputation based two-stage architecture is proposed, where the first stage uses a classifier-guided conditional diffusion component to detect and rectify outliers, and the second stage uses a measurement imputation diffusion component to synthesize missing data points or snippets.
	\item Both stages are based on an enhanced DDIM model with low-length subsequences, which adopts the Bayesian theorem to estimate the precise mean and optimal variance to accelerate the data generation process.
	\item The proposed TSDM is built based on the well-established measurement and system uncertainty model. Consequently, TSDM demonstrates superiority both in recovering SCADA measurements of power systems with renewable energy integration and WAMS measurements with highly nonlinear dynamics under complex power system contingencies.
	\end{enumerate}

The rest of this paper is organized as follows. Problem formulation is presented in Section \ref{sec:Problem}. Section \ref{sec:DDM} briefly introduces the denoising diffusion models and classifier guidance. Section \ref{sec:Improved} elaborates on the design of the proposed TSDM. Two aspects of performance demonstrations are given in Section \ref{sec:CS}. Section \ref{sec:Conclusion} concludes the paper. Supporting lemmas are included in the Appendix for reference.

% section introduction (end)

\section{Problem Formulation} %(fold)
\label{sec:Problem}
In this section, power system state-space model, measurement uncertainty model, and system uncertainty model are introduced. Ultimately, the power system measurement recovery task and challenges are presented in detail.

\subsection{System Model and Physical Constraints}
Power systems usually operate in a quasi-steady state with slow and smooth changes in load or renewable energy generation. However, they transit to a transient state when faced with abrupt contingencies \cite{zhao2019power}. Generally, the SCADA system can effectively monitor the slow oscillations in the quasi-steady state. In contrast, the power system in the transient state requires the utilization of PMUs with higher reporting rates to record dynamic processes. To precisely describe and capture these steady and transient states, the generalized time discretization space model of the power system can be represented as 
\begin{equation} \label{stateeq}
	\begin{aligned}
		\boldsymbol{s}_k=f\left (\boldsymbol{s}_{k-1},\boldsymbol{b}_{k-1},\boldsymbol{u}_k,\boldsymbol{p}  \right ) +\boldsymbol{w}_k	
	\end{aligned}
\end{equation}
\begin{equation} \label{measurementeq}
	\begin{aligned}
		\boldsymbol{z}_k = h(\boldsymbol{s}_k,\boldsymbol{u}_k,\boldsymbol{p})+\boldsymbol{v}_k,	
	\end{aligned}
\end{equation}
where $\boldsymbol{s}_{k-1}$ and $\boldsymbol{s}_k \in\mathbb{R}^{S} $ are the power system state vectors at discrete timesteps $k-1$ and $k$, $\boldsymbol{z}_k \in \mathbb{R}^M$ is the measurement vector containing SCADA and PMU measurements, $\boldsymbol{b}_{k-1}$ is the algebraic state vector, $\boldsymbol{u}_k$ is the system control and input vector, $\boldsymbol{p}$ is the model parameter, $f\left (. \right )$ and $h\left (. \right )$ are nonlinear functions, and $\boldsymbol{w}_k$ and $\boldsymbol{v}_k$ are the state and measurement errors subject to Normal distribution of zero mean and variance matrices of $\boldsymbol{Q}_k$ and $\boldsymbol{R}_k$, respectively. Eq. \eqref{stateeq} is the state-transition equation and Eq. \eqref{measurementeq} is the nonlinear observation equation.

Concretely, assume the power grid is with $m$ generators and $n$ nodes, and the differential algebraic equations (DAEs) of the $i$-th generator are
\begin{equation} \label{dynamice}
	\begin{aligned}
          &\dot{\delta}_{Gi} = \Delta \omega_{Gi}, 2H_{Gi}\Delta \dot{\omega}_{Gi} = P_{mGi} - P_{Gi} - D_{Gi}\Delta \omega_{Gi}\\
          &T'_{qGi0}\dot{E}'_{dGi} = -E'_{dGi} - \left ( X_{qGi}-X'_{qGi} \right )I_{qGi}\\
          &T'_{dGi0}\dot{E}'_{qGi} = E_{fdGi}- E'_{qGi} - \left ( X_{dGi}-X'_{dGi} \right )I_{dGi},
	\end{aligned}
\end{equation}
where $\delta_{Gi}$ is the power angle, $\omega_{Gi}$ is the rotor speed, $H_{Gi}$ is the inertia constant, $P_{mGi}$ is the mechanical input power, $D_{Gi}$ is the damping coefficient, $T'_{qGi0}$ and $T'_{dGi0}$ are the d-axis and q-axis transient time constants, $E'_{dGi}$ and $E'_{qGi}$ are the d-axis and q-axis transient voltages, $E_{fdGi}$ is the field voltage, $X_{dGi}$ and $X_{qGi}$ are d-axis and q-axis synchronous reactances,  $X'_{dGi}$ and $X'_{qGi}$ are d-axis and q-axis transient reactances, $U_{dGi}/U_{qGi}$ and $I_{dGi}/I_{qGi}$ are the d-axis and q-axis voltage and current respectively. WAMS measurements from the PMU installed at the generator terminal bus include generator active/reactive power output $P_{Gi}/Q_{Gi}$, voltage/current magnitude $U_i/I_i$ and phase angle $\theta_i/\gamma_i$.

The measurement equation constraints of the $i$-th generator can be described as
\begin{equation} \label{gcons}
	\begin{aligned}
&\begin{bmatrix}
U_{dGi} \\
U_{qGi} 
\end{bmatrix} = \begin{bmatrix}
\sin(\delta_{Gi}) & -\cos(\delta_{Gi})\\ 
 \cos(\delta_{Gi})& \sin(\delta_{Gi})
\end{bmatrix}\begin{bmatrix}
U_i\cos(\theta_i)\\ 
U_i\sin(\theta_i)
\end{bmatrix} \\
&U_{dGi} = E'_{dGi}+I_{qGi}X'_{qGi}-R_{aGi}I_{dGi}\\
&U_{qGi} = E'_{qGi}-I_{dGi}X'_{dGi}-R_{aGi}I_{qGi}\\
&P_{Gi} = U_{dGi}I_{dGi} + U_{qGi}I_{qGi}, Q_{Gi} = U_{qGi}I_{dGi} - U_{dGi}I_{qGi}\\
&I^2_i = I^2_{dGi}+I^2_{qGi}, \gamma_i = \tan^{-1}\left ( -I_{dGi}/ I_{qGi} \right )+\delta_{Gi},
	\end{aligned}
\end{equation}
where $R_a$ is the stator series resistance. Moreover, the network measurement equation constraints at the $i$-th bus should also be addressed as
\begin{equation} \label{ncons}
	\begin{aligned}
		&P_i = U_i\sum_{j\in \Re_i }U_j\left ( G_{ij}\cos \theta_{ij}+ B_{ij}\sin \theta_{ij} \right ) \\
		&Q_i = U_i\sum_{j\in \Re_i }U_j\left ( G_{ij}\sin \theta_{ij}- B_{ij}\cos \theta_{ij} \right ) \\
		&P_{ij} = U_i^2(g_{ij}+g_{i0}) - U_i U_j \left ( g_{ij}\cos \theta_{ij}+b_{ij}\sin \theta_{ij} \right ) \\ 
		&Q_{ij} = -U_i^2(b_{ij}+b_{i0}) + U_i U_j \left ( b_{ij}\cos \theta_{ij}- g_{ij}\sin \theta_{ij} \right ),
	\end{aligned}
\end{equation}
where $\theta_{ij}$ is the phase angle difference between Buses $i$ and $j$, $G_{ij}+B_{ij}$ is the $ij$-th element of the admittance matrix, $g_{ij}+b_{ij}$ is the admittance of the line connecting Buses $i$ and $j$, $g_{i0}+b_{i0}$ is the admittance of the shunt branch of Bus $i$, and $\Re_i$ represents the bus indexes that are connected to Bus $i$. SCADA measurements include active/reactive power injection $P_i/Q_i$, active/reactive power flow $P_{ij}/Q_{ij}$, and bus magnitude $U_i$.

\begin{figure}[!h]
	\vspace{-0.2cm}
	\centerline{\includegraphics[width=0.48\textwidth]{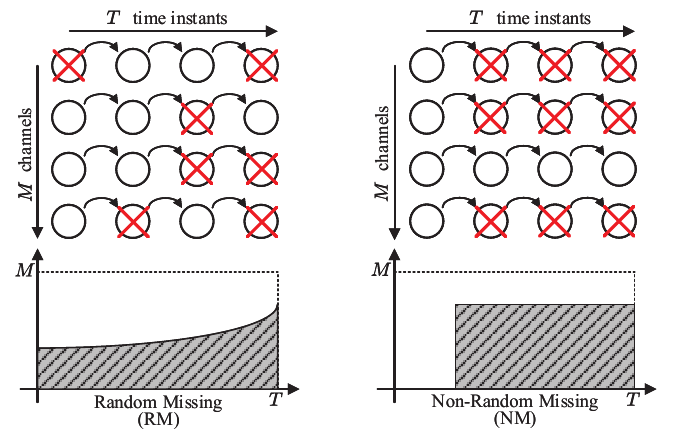}}
	\caption{Random and non-random data losses distribution of received measurements in power systems.}
	\label{datalosses}
	\vspace{-0.2cm}
\end{figure}

\subsection{Measurement Uncertainty Model}
\label{DL}
The observation equation Eq. \eqref{measurementeq} is usually utilized to finish the bad data detection (BDD) tasks in SSE and DSE by computing the residual vector $\boldsymbol{r}_k = \boldsymbol{z}_k - h\left ( \boldsymbol{s}_k \right )$. The elements in $\boldsymbol{r}_k$ indicate whether there are outliers in the corresponding measurements. For this reason, hackers covertly inject a false data component $\Delta \boldsymbol{z}_k^a$ into the measurement vector $\boldsymbol{z}_k$ to offset the state vector $\boldsymbol{s}_k$ with a small enough $\boldsymbol{r}_k^a$ \cite{liu2022false}. Regardless of the control vector and parameter changes, the tampered residual vector can be rewritten as 
\begin{equation} \label{tamperedeq}
	\begin{aligned}
\boldsymbol{r}_k^a = \Delta \boldsymbol{z}_k^a - \left [ h\left ( \boldsymbol{s}_k+\Delta \boldsymbol{s}_k^a \right ) - h\left ( \boldsymbol{s}_k \right ) \right ].
	\end{aligned}
\end{equation}
Evidently, the false data component $\Delta \boldsymbol{z}_k^a$ should be designed to minimize the residual vector $\boldsymbol{r}^a_k$. In this way, attackers can construct false data $\boldsymbol{z}_k^a$ based on state vector offset $\Delta\boldsymbol{s}_k^a$. The general AC nonlinear FDIA model of SCADA and PMU data with incomplete network \cite{liang2016review} information can be described as
\begin{equation} 
	\begin{aligned}
		\mathrm{obj:} \,  &  \mathop{ \mathrm{minimize} }_{\Delta (\boldsymbol{z}_k^a)^{\mho}} \| \underbrace{ \Delta (\boldsymbol{z}_k^a)^{\mho} - \left [ h^{\mho}\left (  (\boldsymbol{s}_k^a)^{\mho} \right ) - h^{\mho}\left ( \boldsymbol{s}^{\mho}_k \right ) \right ]}_{(\boldsymbol{r}_k^a)^{\mho}}  \|_0
 \\
         \mathrm{s.t.:} \,  & (\boldsymbol{s}_k^a)^{\mho} = f^{\mho}\left ( (\boldsymbol{s}_{k-1}^a)^{\mho} \right )+\boldsymbol{w}_k^{\mho},\left | \theta_i^a - \theta_j^a \right | \le \left | \theta_i - \theta_j \right |_{\max}     \\
         &\boldsymbol{s}_{\mathop{\min}}^{\mho}\le (\boldsymbol{s}_k^a)^{\mho}\le \boldsymbol{s}_{\mathop{\max}}^{\mho}, \left | \delta_{Gi}^a - \delta_{Gj}^a \right | \le \left | \delta_{Gi} - \delta_{Gj} \right |_{\max} \\
         &P_{Gi}^{\min}\le P_{Gi}^a \le P_{Gi}^{\max}, Q_{Gi}^{\min}\le Q_{Gi}^a \le Q_{Gi}^{\max},
	\end{aligned}
\end{equation}
where $\mho$ represents the power system area under cyber attacks. Through the above false data construction method and physical constraints, hackers only need to master incomplete power grid information $h^{\mho}(.)$ and $f^{\mho}(.)$ to inject false data that will not be recognized by BDD. In addition, from the time-series perspective, this paper assumes that hackers can launch six temporal forms of DMA, i.e., step, ramp, random noises, replay, phase shift, and amplitude scaling attacks.

Measurements may be continuously missing due to communication malfunction, resulting in multi-channel non-random data losses. Moreover, random missing entries occur frequently as traditional bad data suppression and transmission delay exist in power systems. Anomaly detection in cyber power systems can also result in data losses after removing contaminated data caused by data manipulations. Unfortunately, when a large number of interlocking complex cyber attacks or communication malfunctions occur, the power system may become unobservable and uncontrollable. Collectively, this paper considers two forms of data loss, i.e., random missing (RM) and non-random missing (NM) data. As depicted in Fig. \ref{datalosses}, the received raw SCADA and PMU data consist of measurements from $M$ measurement channels at $T$ time instants. The proportion of RM data is gamma distributed over time \cite{8674594}, and missing entries are randomly distributed in various measurement channels. On the contrary, the distribution of NM data is more regular, and NM usually results in part of $\bm{s}_k$ not being unobservable \cite{zheng2021observability} by conducting sequential data losses at specific measurement channels

\subsection{System Uncertainty Model} \label{SUM}
Denote by $\bm{s}^{*}$ the steady-state values of $\bm{s}$ and define state fluctuation as $\Delta \bm{s} = \bm{s}-\bm{s}^{*}$. The uncertainty of state fluctuation $\Delta \bm{s}$ can be discussed by generator dynamics and network constraints. 

\textit{1) Generator dynamics:} Define a state change vector of the $i$-th generator as $\Delta \bm{s}_{Gi} = [\Delta \delta_{Gi},\Delta \omega_{Gi}, \Delta E'_{dGi}, \Delta E'_{qGi}]^{\top}$. By linearizing Eq. \eqref{dynamice} and considering generator constraints in Eq. \eqref{gcons}, Eq. \eqref{dynamice} becomes
\begin{equation} 
	\begin{aligned}
\Delta \dot{\bm{s}}_{Gi} = \bm{A}_{Gi}\Delta \bm{s}_{Gi}+\bm{B}_{Gi}[\Delta P_{Gi}, \Delta Q_{Gi}]^{\top} + \bm{C}_{Gi}\Delta \bm{u}_{Gi},
	\end{aligned}
\end{equation}	
where $\Delta \bm{u}_{Gi}=[\Delta P_{mGi}, \Delta E_{fdGi}]^{\top}$, and matrices $\bm{A}_{Gi}$, $\bm{B}_{Gi}$, and $\bm{C}_{Gi}$ can be found in Appendix A.

In this way, the dynamics of $m$ generators can be described by 
\begin{equation} \label{gdr}
	\begin{aligned}
\Delta \dot{\bm{s}}_{G} = \bm{A}_{G}\Delta \bm{s}_{G}+\bm{B}_{G}\Delta \bm{PQ}_{G} + \bm{C}_{G}\Delta \bm{u}_{G},
\end{aligned} 
    \end{equation}
where $\Delta \bm{PQ}_G = [\Delta P_{G1}, \Delta Q_{G1}, \cdots, \Delta P_{Gm}, \Delta Q_{Gm}]^{\top}$, $\bm{A}_G = diag(\bm{A}_{G1}\cdots \bm{A}_{Gm})$, $\bm{B}_G = diag(\bm{B}_{G1}\cdots \bm{B}_{Gm})$, $\bm{C}_G = diag(\bm{C}_{G1}\cdots \bm{C}_{Gm})$, $\Delta \bm{s}_G=[\Delta s_{G1}^{\top},\cdots, \Delta s_{Gm}^{\top}]^{\top}$, $\Delta \bm{u}_{G} = [\Delta u_{G1}^{\top},\cdots, \Delta u_{Gm}^{\top}]^{\top}$. After discretization, Eq. \eqref{gdr} becomes
\begin{equation} \label{dgdr}
	\begin{aligned}
		\Delta \bm{s}_{G,k} = & (1+\Delta t\bm{A}_G)\Delta \bm{s}_{G,k-1}\\
		& +\Delta t\bm{B}_G\Delta \bm{PQ}_{G,k}+\Delta t\bm{C}_G\Delta \bm{u}_{G,k},
	\end{aligned} 
\end{equation}
where $\Delta t$ is the sampling interval.

\textit{2) Network constraints:} Power system should be constrainted by nonlinear power injection in Eq. \eqref{ncons}. Correspondingly, linearize the $i$-th bus's power injection equation at stable equilibrium point \cite{machowski2020power, wei2023cyber}, it becomes
\begin{equation} \label{pjel}
	\begin{aligned}
		&\Delta P_i =\Delta P_{Gi}-\Delta P_{Li}= \sum_{j=1}^{n}H_{ij}\Delta \theta_j  + \sum_{j=1}^{n}N_{ij}\Delta U_j \\
		&\Delta Q_i = \Delta Q_{Gi}-\Delta Q_{Li}= \sum_{j=1}^{n}M_{ij}\Delta \theta_j  + \sum_{j=1}^{n}L_{ij}\Delta U_j,
	\end{aligned} 
\end{equation}
where $\Delta P_{Li}$ and $\Delta Q_{Li}$ are the power injection by the load at the $i$-th bus, $\bm{H}$, $\bm{N}$, $\bm{M}$, and $\bm{L}$ are real-value matrices during the linearization. Eq. \eqref{pjel} can be rewritten in matrix form as
\begin{equation} \label{ncr}
	\begin{aligned}
		\begin{bmatrix}
			\Delta \bm{P}_G - \Delta \bm{P}_L\\ 
			\Delta \bm{Q}_G - \Delta \bm{Q}_L
			\end{bmatrix} = \underbrace{\begin{bmatrix}
			\bm{H}_{n\times n} & \bm{N}_{n\times n}\\ 
			 \bm{M}_{n\times n}& \bm{L}_{n\times n}
			\end{bmatrix}}_{\bm{D}'_{2n\times 2n}} \begin{bmatrix}
			\Delta \bm{\theta_n}\\ 
			\Delta \bm{U_n}
			\end{bmatrix}.
	\end{aligned} 
\end{equation}	
The bus voltage phasors fluctuation can be denoted by $[\Delta \bm{\theta}^{\top},\Delta \bm{U}^{\top}]^{\top}=\bm{D}'^{-1}\left ( [\Delta \bm{P}_G^{\top},\Delta \bm{Q}_G^{\top}]^{\top} \! - [\Delta \bm{P}_L^{\top},\Delta \bm{Q}_L^{\top}]^{\top}\right )$.

\textit{3) Uncertainty model:} The power system state uncertainty can be defined as $\Delta \bm{s} = [\Delta \bm{s}^{\top}_G, \Delta \bm{\theta}^{\top}, \Delta \bm{U}^{\top}]^{\top}$. As can be seen in Eq. \eqref{dgdr} and Eq. \eqref{ncr}, $\Delta \bm{s} $ is bounded by current system state, generator power output, load demand, and system control input by 
\begin{equation} \label{ume}
	\begin{aligned}
\Delta \bm{s}_k = & \bm{A} \Delta \bm{s}_{k-1} + \bm{B}[\bm{P}_{G,k}^{\top},\Delta \bm{Q}_{G,k}^{\top}]^{\top} \\
& +\bm{C} \Delta \bm{u}_k - \bm{D} [\bm{P}_{L,k}^{\top},\Delta \bm{Q}_{L,k}^{\top}]^{\top},
	\end{aligned} 
\end{equation}
where matrices $\bm{A}$, $\bm{B}$, $\bm{C}$, and $\bm{D}$ are defined in Appendix A. For simplicity, assume the fluctuations of active and reactive power are similar, a function $\Delta \bm{s}_k = \mathcal{F}(\Delta \bm{s}_{k-1}, \Delta \bm{P}_{G,k},\Delta \bm{P}_{L,k}, \Delta \bm{u}_k)$ denotes the uncertainty model. The power output of renewables brings uncertainties to $\Delta \bm{P}_G$, and sporadic various events increase the complexity of the input vector $\Delta \bm{u}$. As a consequence, the system will exhibit strong randomness and highly nonlinear dynamics.

\subsection{Measurement Recovery Task and Challenges}
Assume that FDIA and data loss do not happen simultaneously, and the measurement vector $\bm{z}_k$ sampled at time $k$ is contaminated by FDIA or data loss and becomes $\bm{z}_k^a$. The goal of power system measurement recovery task is to find the original $\bm{z}_k$ given $\bm{z}_k^a$ by the mapping distribution $q(\bm{z}_k|\bm{z}_k^a)$, with an optimization objective
\begin{equation} \label{mro}
	\begin{aligned}
		\mathop{\arg\max}_{\bm{z}_k}\log q(\bm{z}_k| \bm{z}^a_k) = \mathop{\arg\max}_{\bm{z}_k}[\log q(\bm{z}^a_k|\bm{z}_k)+ \log q(\bm{z}_k)].
	\end{aligned} 
\end{equation}	
However, it is difficult to achieve high accuracy by utilizing only a single snapshot of measurements for recovery. As power system measurements have demonstrated strong spatial-temporal correlation \cite{zhang2021spatio} and low-rank property \cite{7080935}, the measurement matrix 
\begin{equation} 
	\begin{aligned}	
\bm{y}_0 = [(\bm{z}^a_{k-T+1})^{\top},(\bm{z}^a_{k-T+2})^{\top}, \cdots,(\bm{z}^a_{k})^{\top} ]^{\top} \in \mathbb{R}^{M\times T}
	\end{aligned} 
\end{equation}
that consists of contaminated $\bm{z}^a$ over $T$ time intervals is used, and $\bm{x}_0$ represents the original data. The objective Eq. \eqref{mro} can be rewritten as
\begin{equation} \label{mroi}
	\begin{aligned}
		\mathop{\arg\max}_{\bm{x}_0}\left [\underbrace{\log q(\bm{y}_0|\bm{x}_0)}_{\textrm{measurement consistency term}}+ \underbrace{\log q(\bm{x}_0)}_{\textrm{measurement prior term}} \right ].
	\end{aligned} 
\end{equation}	
To this end, modern power system data recovery task Eq. \eqref{mroi} faces three main challenges:
\begin{enumerate}
	\item According to Subsection \ref{DL}, since the measurement corruption type and severity are entirely unknown and uncertain, the performance of supervised learning models that directly map $\bm{y}_0$ to $\bm{x}_0$ will be questioned.
	\item According to Subsection \ref{SUM}, a model with powerful generation ability should be established to model the complex system uncertainty as a prior knowledge $q(\bm{x}_0)$. 
	\item The measurement consistency term in Eq. \eqref{mroi} must be considered to quickly find the optimal solution for the current observations.
	\end{enumerate}

\begin{figure}[!h]
	\vspace{-0.4cm}
	\centerline{\includegraphics[width=0.48\textwidth]{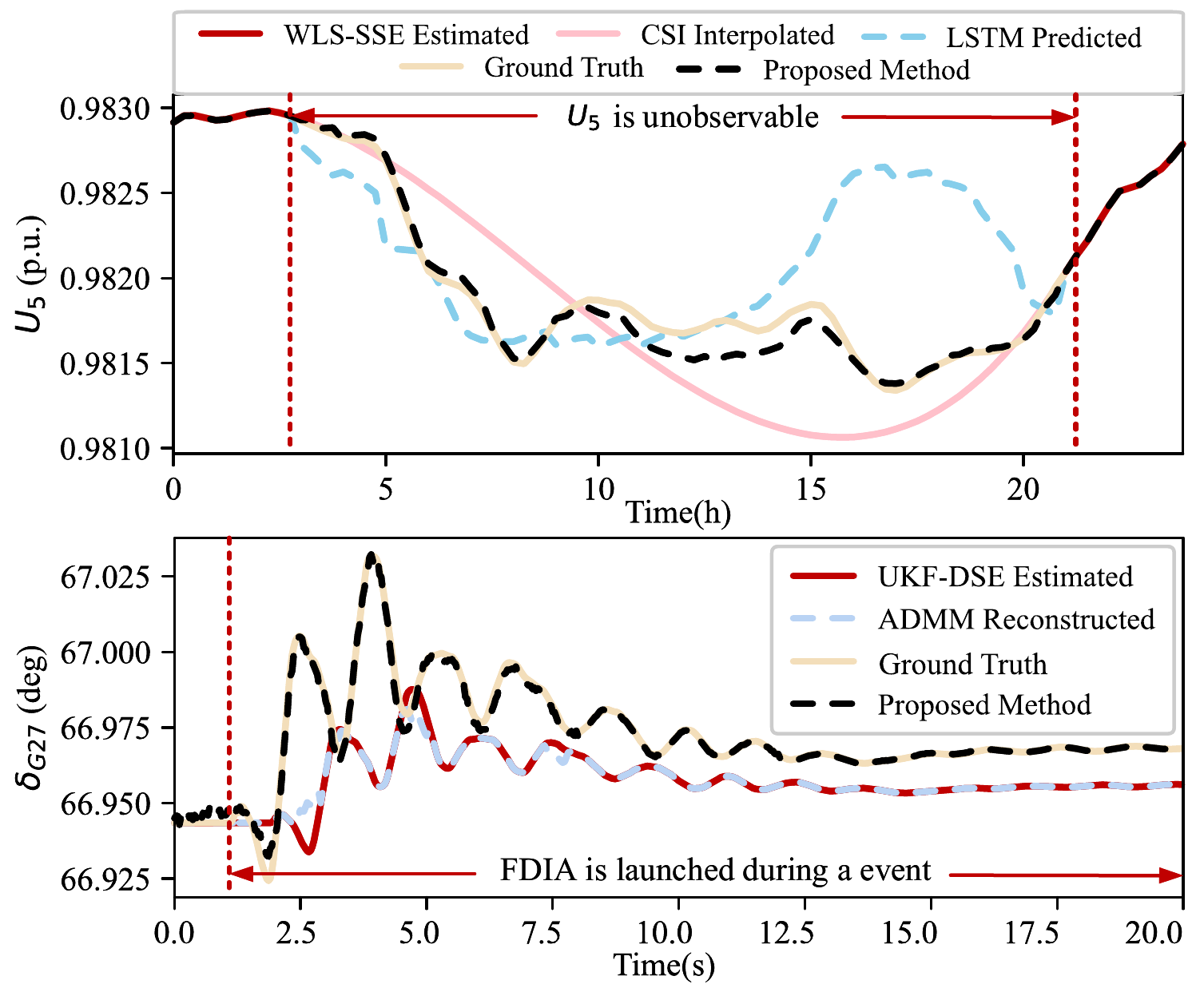}}
	\caption{Measurement recovery challenges under irregular $\Delta \bm{P}_G$ and nonlinear dynamics by $\Delta \bm{u}$.}
	\label{challenges1}
	\vspace{-0.4cm}
\end{figure}

Two cases are used to demonstrate the cyber-physical challenges. In Fig. \ref{challenges1}, the voltage magnitude $U_5$ of Bus 5 in the state vector $\bm{s}$ of IEEE 30-bus system is unobservable due to data loss in WLS-based SSE. The classic cubic spline interpolation (CSI) \cite{2020An} and LSTM temporal prediction \cite{8356714} are used for measurement completion. The results show that due to the irregular $\Delta \bm{P}_G$ of renewables, those recovery methods can not address this challenge. Fig. \ref{challenges1} shows a case of highly nonlinear DSE approach, where the classic UKF \cite{qi2016dynamic} tries to track the power angle $\delta_{G27}$ of the Generator 27 of NPCC 140-bus system.  However, due to the presence of maliciously injected false data, the estimated $\delta_{G27}$ of UKF gradually deviates from real values after 1s. RPCA based on ADMM algorithm \cite{8120129} is also used to filter out these outliers with low accuracy under the premise of high nonlinearity and complex $\Delta \bm{u}$.	

% section Problem (end)

\section{Background: Denoising Diffusion Models} %(fold)
\label{sec:DDM}
Diffusion models can be selected to model the system uncertainty as they have led the artificial intelligence-generated content (AIGC) field. In this section, the basics of DDPM and DDIM are briefly introduced, and Subsection \ref{calssifierguidance} presents the approach to maximize the measurement consistency term.

\subsection{Denoising Diffusion Probabilistic Models}
Denoising diffusion models are a class of machine learning algorithms inspired by Non-Equilibrium Thermodynamics that have better performance than generative models such as GANs. Diffusion models generally include a forward diffusion process and a reverse prediction process. The forward process of DDPM \cite{2020DDPM} refers to gradually adding Gaussian noises to the original measurements $\boldsymbol{x}_0 \sim q(\boldsymbol{x}_0)$ until the measurements become random noises $\boldsymbol{x}_N \sim q(\boldsymbol{x}_N | \boldsymbol{x}_0)$ normally distributed with zero mean and unit variance. The forward process with $N$ steps can be described as 
\begin{equation}
	\begin{aligned}
		&q(\boldsymbol{x}_{1:N} |\boldsymbol{x}_0) = \prod_{n=1}^{N} q(\boldsymbol{x}_{n} |\boldsymbol{x}_{n-1}) \\
		&q(\boldsymbol{x}_{n} |\boldsymbol{x}_{n-1})=\mathcal{N}(\bm{x}_n; \sqrt{1-\beta_n}\boldsymbol{x}_{n-1},\beta_n\boldsymbol{I}),		
	\end{aligned}
\end{equation}
where $\left \{ \beta_n  \right \} ^N_{n=1}$ is the variance used for each step and $\beta_n \in (0,1)$, and the variance schedule satisfies the relationship $\beta_1<\beta_2<\dots <\beta_N$. The forward process is a Markov chain, and $\boldsymbol{x}_n$ at any step can be computed as
\begin{equation}
	\begin{aligned}
		q(\boldsymbol{x}_n |\boldsymbol{x}_0) = \mathcal{N}(\bm{x}_n; \sqrt{\alpha_n} \boldsymbol{x}_0,(1-\alpha_n)\boldsymbol{I}),		
	\end{aligned}
\end{equation}
where $\alpha_n =  {\textstyle \prod_{i=1}^{n}}(1-\beta_i) $. As the number of diffusion steps increases, the signal ratio $\sqrt{\alpha_n}$ will gradually approach 0, while the noise ratio $\sqrt{1-\alpha_n}$ will approach 1, ensuring that the final $\boldsymbol{x}_N$ is close to a random noise $\epsilon\sim \mathcal{N}(\bm{0}, \boldsymbol{I})$.

The reverse process of DDPM is opposite to the forward process in terms of noise elimination, i.e., the reverse process generates $\boldsymbol{x}_0$ from the random noise $\boldsymbol{x}_N$. In order to estimate the distribution $q(\boldsymbol{x}_{n-1}|\boldsymbol{x}_n)$, DDPM also defines the reverse process as a Markov chain and utilizes the neural network parameterized function $p_{\theta}(\boldsymbol{x}_{n-1}|\boldsymbol{x}_n)=\mathcal{N}\left ( \bm{x}_{n-1}; \mu_{\theta}(\boldsymbol{x}_n,n),\Sigma_{\theta}(\boldsymbol{x}_n,n) \right )$ to predict the mean $\mu_{\theta}$ and variance $\Sigma_{\theta}$ from $\boldsymbol{x}_n$ at step $n$ of the reverse process. DDPM ultimately obtains the training objective $L_{simple} = \mathbb{E}_{\boldsymbol{x}_0, \epsilon \sim \mathcal{N}(\bm{0},\boldsymbol{I})}\left [ \left \|\epsilon -\epsilon_{\theta}(\boldsymbol{x}_n,n)   \right \|^2  \right ]$ by considering the variational lower bound, where $\epsilon_{\theta}$ is a neural network intended to predict the actual noise $\epsilon$ from $\boldsymbol{x}_n$.

\subsection{Denoising Diffusion Implicit Models}
Building upon DDPM, DDIM \cite{2020Denoising} optimizes and accelerates the diffusion models by considering a non-Markov chain derivation. Since DDIM does not define a forward process, the target distribution of the reverse process  
\begin{equation}
	\begin{aligned}
q\left ( \boldsymbol{x}_{n-1}| \boldsymbol{x}_{n},\boldsymbol{x}_{0} \right ) = \frac{q\left ( \boldsymbol{x}_{n}| \boldsymbol{x}_{n-1} \right ) q\left ( \boldsymbol{x}_{n-1}| \boldsymbol{x}_{0} \right ) }{q\left ( \boldsymbol{x}_{n}| \boldsymbol{x}_{0} \right ) }	
	\end{aligned}
\end{equation}
should be obtained, considering that $q(\boldsymbol{x}_n | \boldsymbol{x}_{n-1})$ is not given. As a consequence, the derivation of the reverse process  only needs to meet the marginal distribution condition. It can be inferred in \cite{2020Denoising} that the target distribution with variable standard deviation parameter $\sigma_n$ for all $n\ge 2$ is
\begin{equation}
	\begin{aligned}
		&q_{\sigma_n}(\boldsymbol{x}_{n-1}|\boldsymbol{x}_n,\boldsymbol{x}_0) = \\
		&\mathcal{N} (\bm{x}_{n-1}; \sqrt{\alpha_{n-1} }\boldsymbol{x}_0+\sqrt{1-\alpha_{n-1}-\sigma^2_n  }\frac{\boldsymbol{x}_n-\sqrt{\alpha_n}\boldsymbol{x}_0}{\sqrt{1-\alpha_n }},\sigma_n^2\boldsymbol{I}  ).	
	\end{aligned}
\end{equation}
Since $\boldsymbol{x}_0$ is unknown, the estimation $\bar{\mu}(\boldsymbol{x}_n)$ of $\boldsymbol{x}_0$ can be derived from $\boldsymbol{x}_n=\sqrt{\alpha_n}\boldsymbol{x}_0+\sqrt{1-\alpha_n}\epsilon $ as 
\begin{equation} \label{meanofx0}
	\begin{aligned}
		\bar{\mu}(\boldsymbol{x}_n) = \frac{1}{\sqrt{\alpha_n }}  \left [ \boldsymbol{x}_n - \sqrt{1-\alpha_n}\epsilon_{\theta}(\boldsymbol{x}_n,n)  \right ].	
	\end{aligned}
\end{equation}
Accordingly, during the generation phase, $\boldsymbol{x}_{n-1}$ can be computed by the distribution $q_{\sigma_n}$, where the variable $\sigma_n$ does not have any constraints. When $\sigma_n = \sqrt{(1-\alpha_{n-1} )/(1-\alpha_n )}\sqrt{1-\alpha_n/\alpha_{n-1}}$, the diffusion model becomes DDPM. When $\sigma_n = 0$, the reverse process of the diffusion model is deterministic, and the input random noise will generate fixed target measurements: this type of diffusion model is called DDIM.

\subsection{Classifier-Guidance} \label{calssifierguidance}
Compared to other generative models, diffusion models have stronger potential editable and controllable properties with multi-step sampling process by classifier-free \cite{ho2022classifier} and classifier-guidance \cite{dhariwal2021diffusion} approaches. Among them, classifier-guidance-based diffusion model can generate target distributions without additional learning process. In this way, diffusion models can be controlled to output corresponding restored power system measurements according to the input condition $\boldsymbol{y}$, i.e., under the control of the corrupted matrix $\boldsymbol{y}_0$. In the reverse process, the most critical step is the construction of distribution $p_{\theta}(\boldsymbol{x}_{n-1}|\boldsymbol{x}_n)$. When the condition $\boldsymbol{y}$ is added into the input, according to Bayes's theorem, the conditioned $p_{\theta}$ can be described as 
\begin{equation} \label{exponential}
	\begin{aligned}
		& p_{\theta}(\boldsymbol{x}_{n-1}|\boldsymbol{x}_n,\boldsymbol{y})  =\frac{p_{\theta}(\boldsymbol{x}_{n-1}|\boldsymbol{x}_n)p_{\theta}(\boldsymbol{y}|\boldsymbol{x}_{n-1},\boldsymbol{x}_n)}{p_{\theta}(\boldsymbol{y}|\boldsymbol{x}_n)} \\
		& \approx p_{\theta}(\boldsymbol{x}_{n-1}|\boldsymbol{x}_n)e^{\log{p_{\theta}(\boldsymbol{y}|\boldsymbol{x}_{n-1})}-\log{p_{\theta}(\boldsymbol{y}|\boldsymbol{x}_{n})} }.	
	\end{aligned}
\end{equation}
After Taylor expansion, the exponential term in Eq. \eqref{exponential} becomes 
\begin{equation}
	\begin{aligned}
		\log\left [ {p_{\theta}(\boldsymbol{y}|\boldsymbol{x}_{n-1})}/{p_{\theta}(\boldsymbol{y}|\boldsymbol{x}_{n})} \right ] \approx (\boldsymbol{x}_{n-1}-\boldsymbol{x}_{n})\nabla_{\boldsymbol{x}_{n}}\log{p_{\theta}(\boldsymbol{y}|\boldsymbol{x}_n)}.
	\end{aligned}
\end{equation}
Consequently, the conditioned diffusion models with classifier guidance can approximately utilize the gradient $\nabla_{\boldsymbol{x}_{n}}\log{p_{\theta}(\boldsymbol{y}|\boldsymbol{x}_n)}$ to guide the random noises to gradually approach the target distribution. The gradient of the reverse process is $\nabla_{\boldsymbol{x}_{n}}\log{p_{\theta}(\boldsymbol{x}_n|\boldsymbol{y}})$, which can be transformed into $\nabla_{\boldsymbol{x}_{n}}\log{p_{\theta}(\boldsymbol{x}_n)}  +\nabla_{\boldsymbol{x}_{n}}\log{p_{\theta}(\boldsymbol{y}|\boldsymbol{x}_n)}$ by Bayes's theorem. It can be seen from the score matching method \cite{daras2022soft} that when the diffusion models predict noise $\epsilon_{\theta}$, $\nabla_{\boldsymbol{x}_{n}}\log{p_{\theta}(\boldsymbol{x}_n)}$ equals $-\epsilon_{\theta}(\boldsymbol{x}_n,n)/\sqrt{1-\alpha_n}$, so the gradient can be rewritten as 
\begin{equation} \label{gradient}
	\begin{aligned}
		&\nabla_{\boldsymbol{x}_{n}}\log{p_{\theta}(\boldsymbol{x}_n|\boldsymbol{y}}) \\
           &=-\frac{\epsilon_{\theta}(\boldsymbol{x}_n,n)-\sqrt{1-\alpha_n}\nabla_{\boldsymbol{x}_{n}}\log{p_{\theta}(\boldsymbol{y}|\boldsymbol{x}_n)}}{\sqrt{1-\alpha_n}} .	
	\end{aligned}
\end{equation}
As a result, regardless of the variance of the generation process, the diffusion models only need to replace $\epsilon_{\theta}(\boldsymbol{x}_n,n)$ with $\epsilon_{\theta}(\boldsymbol{x}_n,n)-\sqrt{1-\alpha_n}\nabla_{\boldsymbol{x}_{n}}\log{p_{\theta}(\boldsymbol{y}|\boldsymbol{x}_n)}$ to conditionally control the measurements generation process.

% section DDM (end)

\section{Enhanced Diffusion Recovery Model} %(fold)
\label{sec:Improved}
The training for a power system measurement prior model is firstly introduced in this section. Subsection \ref{SAOV} clarifies the adpoted faster sampling form. Subsequently, controllable measurement generation methods with contaminated or incomplete condition are established by utilizing the same pre-trained prior model and sampling form. Subsection \ref{ITSDM} and \ref{CPITSDM} present the proposed TSDM and its cyber-physical implementation in modern power systems.

\subsection{Diffusion Measurement Prior Model}
\begin{figure}[!h]
	\vspace{-0.2cm}
	\centerline{\includegraphics[width=0.48\textwidth]{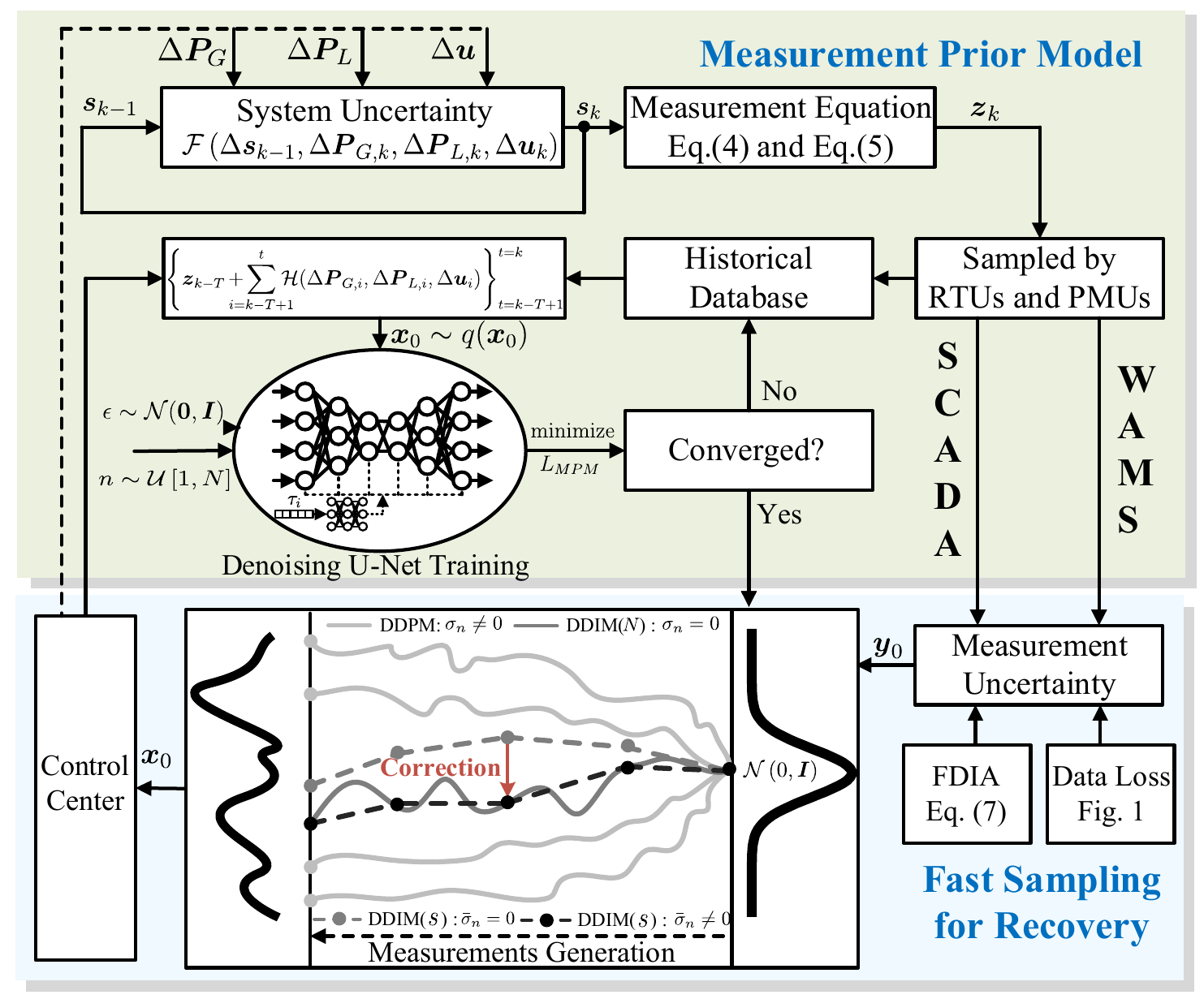}}
	\caption{Diffusion measurement prior model and its fast sampling method.}
	\label{prior_scheme}
	\vspace{-0.2cm}
\end{figure}

According to Subsection \ref{SUM}, the system state fluctuations are the responses to the changes $\Delta \bm{P}_G$, $\Delta \bm{P}_L$, and $\Delta \bm{u}$ by function $\mathcal{F}(.)$. Deployed RTUs and PMUs sample the measurements $\bm{z}_k$ with measurement equations Eq. \eqref{gcons} and Eq. \eqref{ncons} to monitor the system states. To this end, the measurements change can be captured by 
\begin{equation} 
	\begin{aligned}
\Delta \bm{z}_k &= h \left (  \mathcal{F}( \Delta \bm{s}_{k-1}, \Delta \bm{P}_{G,k}, \Delta \bm{P}_{L,k}, \Delta \bm{u}_{k}) \right ) \\
&= \mathcal{H}(\Delta \bm{s}_{k-1}, \Delta \bm{P}_{G,k}, \Delta \bm{P}_{L,k}, \Delta \bm{u}_{k} ),
\end{aligned}
\end{equation}
where $\mathcal{H}(.)$ denotes the mapping function between measurements and system uncertainties. DDPM is implemented based on the denoising U-Net to be trained for a measurement prior model (MPM) to maximize the prior term in Eq. \eqref{mroi}, which is depicted in Fig. \ref{prior_scheme}. It aims to minimize the gap between the prior model $p_{\theta}$ and real observations, i.e., minimizing the MPM loss
\begin{equation} 
    \begin{aligned}
&L_{MPM} =  \mathbb{E}_{\bm{x}_0 \sim q \left ( \left \{ \bm{z}_{k-T} + \sum^t_{i=k-T+1}\Delta \bm{z}_i \right \}_{t=k-T+1}^{t=k}\right ) } \\
& \Bigg [ \underbrace{D_{KL} \left (  q(\bm{x}_N | \bm{x}_0) \parallel p_{\theta}(\bm{x}_N)   \right )}_{L_N} - \underbrace{\log p_{\theta}(\bm{x}_0 | \bm{x}_1 )}_{L_0} \\
& + \sum^N_{n=2} \underbrace{D_{KL} \left (  q(\bm{x}_{n-1} | \bm{x}_n, \bm{x}_0) \parallel p_{\theta}(\bm{x}_{n-1} | \bm{x}_n) \right )}_{L_{n-1}}  \Bigg ], 
\end{aligned}
\end{equation}
where $D_{KL}(.\parallel .)$ is the Kullback-Leibler divergence between two distributions. Evidently, unlike GANs that directly matching generated distribution to the target distribution, the diffusion models divide the optimization objective into $N$ terms with higher training and sampling demands. In this way, with multi-steps, MPM is more likely to capture or characterize the dynamics $\Delta \bm{z}_k$ induced by irregular $\Delta \bm{P}_G$ and complex $\Delta \bm{u}$. 

 The denoising U-Net is an artificial neural network with an autoencoder architecture combined with a residual block and an attention block. The optimization objective of the denoising U-Net is to make the predicted noises $\epsilon_{\theta}$ consistent with the real noises $\epsilon$, i.e., denoising U-Net randomly selects a normal sample $\bm{x}_0 $ and step $n$ during training process, and then calculates the $L_2$ loss between $\epsilon_{\theta}$ and $\epsilon$ to update the gradient and network parameters $\theta$. Moreover, the U-Net encodes step $n$ into the network in a time embedding manner. 

The N-step training and sampling strategy that enhances the generation performance of diffusion models also brings the disadvantage of excessive time and resource consumption, with the number of function evaluations (NFE) reaching $N$. The acceleration methods of diffusion models are very important, as power system data recovery often requires less time consumption and should even be real-time in WAMSs. Various methods based on DDIM \cite{2020Denoising}, optimal variance \cite{bao2022analytic}, ODE solver \cite{zhou2023fast}, knowledge distillation \cite{song2023consistency}, and latent diffusion model \cite{rombach2022high} have effectively improved sampling speed. However, some methods have high implementation costs, so this paper adopts a low-cost acceleration strategy that combines DDIM subsequence sampling with optimal variance.

\subsection{Subsequence Acceleration with Optimal Variance} \label{SAOV}
DDIM sets variance $\sigma_n =0$ to make the generation process of diffusion model become deterministic. In order to accelerate the reverse process of DDIM, the data generation trajectory should be further optimized. Accordingly, the setting of optimal variance is considered as one of the means to accelerate the diffusion model. The target distribution in the reverse process of DDIM can be rewritten as
\begin{equation} \label{DDIMreverse}
	\begin{aligned}
		&p(\boldsymbol{x}_{n-1}|\boldsymbol{x}_n,\boldsymbol{x}_0) =\mathcal{N} (\bm{x}_{n-1}; \frac{\sqrt{1-\alpha_{n-1}-\sigma_n^2  }}{\sqrt{1-\alpha_n}}\boldsymbol{x}_n \\
		&+ \underbrace{( \sqrt{\alpha_{n-1} }-\frac{\sqrt{1-\alpha_{n-1}-\sigma_n^2   }\sqrt{\alpha_n}}{\sqrt{1-\alpha_n}}  )}_{\gamma_n }\boldsymbol{x}_0 ,\sigma_n^2\boldsymbol{I}   ) .	
	\end{aligned}
\end{equation}
In classical DDIM, $\bar{\mu }(\boldsymbol{x}_n) $ estimated by $\boldsymbol{x}_n$ is utilized to replace $\boldsymbol{x}_0$ in Eq. \eqref{DDIMreverse}. Nevertheless, $\bar{\mu }(\boldsymbol{x}_n) $ predicted by $\boldsymbol{x}_n$ can hardly be completely accurate, and the trajectory is uncertain. Consequently, the proposed improved diffusion model utilizes Normal distribution $\mathcal{N}(\bm{x}_0; \bar{\mu}(\boldsymbol{x}_n), \bar{\sigma}_n^2 \boldsymbol{I})$ with mean $\bar{\mu}(\boldsymbol{x}_n)$ and optimal variance $\bar{\sigma}_n^2$ to approximate $\boldsymbol{x}_0$, and the optimal variance (Appendix B) \cite{bao2022analytic} can be expressed as 
\begin{equation} \label{optimalv}
	\begin{aligned}
		\bar{\sigma}_n^2 = \frac{1-\alpha_n }{\alpha_n} \left \{  1- \frac{1}{M \times T} \mathbb{E}_{\boldsymbol{x}_n\sim p(\boldsymbol{x}_n)}\left [ \left \| \epsilon_{\theta}(\boldsymbol{x}_n,n) \right \|^2_2  \right ] \right \}.	
	\end{aligned}
\end{equation}
The optimal variance schedule $\left \{ \bar{\sigma}^2 \right \}^N_{n=1}$ will be calculated based on normal dataset of each power system according to Eq. \eqref{optimalv} and saved offline before data recovery sampling. When the model is used for data recovery, optimal variance $\bar{\sigma}^2_n$ will be directly used for correction in the reverse process.

\begin{figure}[!h]
	\vspace{-0.2cm}
	\centerline{\includegraphics[width=0.48\textwidth]{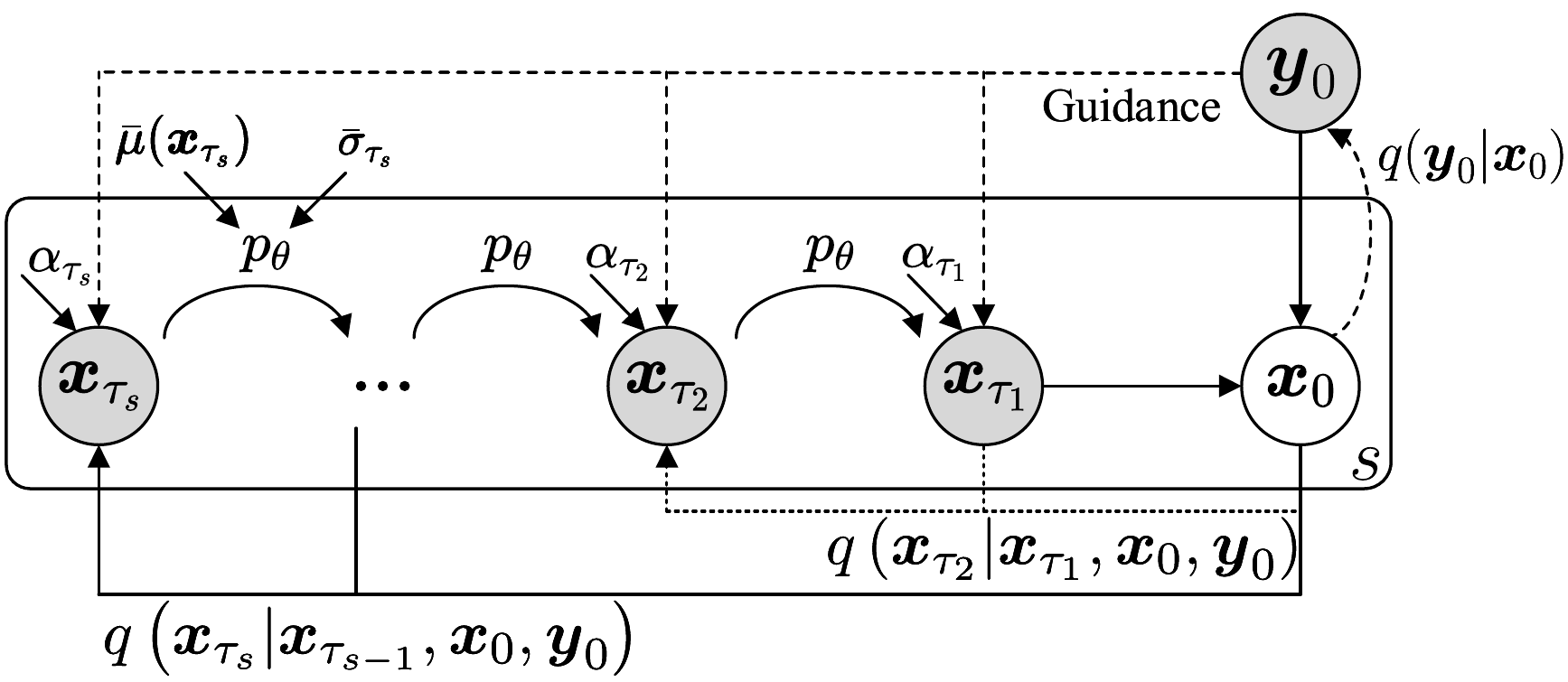}}
	\caption{Improved denoising diffusion implicit model with subsequence acceleration and parameterized reverse process $p_{\theta}$ under classifier-guidance.}
	\label{subsequence}
	\vspace{-0.2cm}
	\end{figure}

Compared with DDPM, DDIM does not clearly define the forward process in derivation, so DDIM can assume a forward process with shorter steps, i.e., DDIM can sample a subsequence $\boldsymbol{\tau}= [\tau_1, \tau_2, \dots, \tau_s]$ with a length of $s$ from the original sequence $[1,2, \dots, N]$. Apparently, DDIM can reuse the diffusion model trained by DDPM, and the training results of DDPM essentially include the training results of its arbitrary subsequence parameter $\boldsymbol{\tau}$. As a result, DDIM can reduce the generation step size from $N$ to $s$, leading to a computational complexity reduction of $s/N$ for power system data recovery. The diffusion model under classifier-guidance accelerated by subsequence is depicted in Fig. \ref{subsequence}.

Each step of the reverse process can be represented by the target distribution $p_{\theta}$ parameterized by the neural network, provided that the estimation $\bar{\mu}(\boldsymbol{x}_{\tau_i})$ of $\boldsymbol{x}$ and optimal variance $\bar{\sigma}^2_{\tau_i}$ need to be calculated previously. Moreover, let 
\begin{equation}
	\begin{aligned}
		\Gamma_{\tau_i}=\sqrt{\alpha_{\tau_{i-1}} }-\frac{\sqrt{1-\alpha_{\tau_{i-1}} }\sqrt{\alpha_{\tau_i}}}{\sqrt{1-\alpha_{\tau_i}}}.	
	\end{aligned}
\end{equation}
Conclusively, the final update equation with subsequence acceleration can be represented as 
\begin{equation} \label{updatedata}
	\begin{aligned}
		\boldsymbol{x}_{\tau_{i-1}}= \frac{\sqrt{1-\alpha_{\tau_{i-1}}}}{\sqrt{1-\alpha_{\tau_i}}} \boldsymbol{x}_{\tau_i}+\Gamma_{\tau_i}\bar{\mu}(\boldsymbol{x}_{\tau_i})+\Gamma_{\tau_i}\bar{\sigma}_{\tau_i}\epsilon.	
	\end{aligned}
\end{equation}
Alternatively, the detailed sampling form of $p_{\theta}$ can be described as
\begin{equation} \label{updatedata2}
	\begin{aligned}
		\boldsymbol{x}_{\tau_{i-1}} = & \frac{\sqrt{\alpha_{\tau_{i-1}}} }{\sqrt{\alpha_{\tau_i}}}\boldsymbol{x}_{\tau_i} +\left ( \sqrt{\alpha_{\tau_{i-1}} }-\frac{\sqrt{1-\alpha_{\tau_{i-1}} }\sqrt{\alpha_{\tau_i} }}{\sqrt{1-\alpha_{\tau_i}}}  \right )\bar{\sigma}_{\tau_i}\epsilon  \\
		&+\left (   \sqrt{1-\alpha_{\tau_{i-1}}}- \frac{\sqrt{\alpha_{\tau_{i-1}} }\sqrt{1-\alpha_{\tau_i} }  }{\sqrt{\alpha_{\tau_i}}}  \right ) \epsilon_{\theta}(\boldsymbol{x}_{\tau_i},\tau_i).	
	\end{aligned}
\end{equation}

\subsection{Conditional Diffusion Model}
\label{conditionaldiffusion}
To maximize the consistency term $\log q(\bm{y}_0 | \bm{x}_0)$, as shown in Fig. \ref{subsequence}, $\bm{y}_0$ should guide the measurement generation. In many applications of diffusion models with classifier guidance, the unconditional score term $\nabla_{\boldsymbol{x}_{\tau_i}}\log p_{\theta}(\boldsymbol{x}_{\tau_i})$ can be dierctly computed as $-\epsilon_{\theta}(\boldsymbol{x}_{\tau_i}, \tau_i) / \sqrt{1-\alpha_{\tau_i}}$, while the calculation of the conditional score $\nabla_{\boldsymbol{x}_{\tau_i}}\log p_{\theta}(  \boldsymbol{y}_{\tau_i} | \boldsymbol{x}_{\tau_i})$ in subsection \ref{calssifierguidance} is very complicated. Intuitively, the conditional score term can be approximated as the difference or correction score between $\boldsymbol{y}_{\tau_i}$ and $\boldsymbol{x}_{\tau_i}$ in each denoising step, where $\boldsymbol{y}_{\tau_i}$ is computed by
\begin{equation}
	\begin{aligned}
		\bm{y}_{\tau_i} = \sqrt{\alpha_{\tau_i}}\boldsymbol{y}_0+ \sqrt{1-\alpha_{\tau_i}}\epsilon_{\theta}(\boldsymbol{x}_{\tau_i},\tau_i).
	\end{aligned}
\end{equation}
As a consequence, $\boldsymbol{y}_{\tau_i}-\boldsymbol{x}_{\tau_i}$ can be utilized as the likelihood to guide the deterministic data generation process. When $\boldsymbol{y}_{\tau_i}$ is obtained by adding noises to the raw measurements $\boldsymbol{y}_0$ contaminated by hackers or communication malfunction, the corresponding $\boldsymbol{x}_0$ generated by the reverse process will have a significant difference from $\boldsymbol{y}_0$. As a result, the discrepancies can be considered as corrupted measurements, thus realizing the localization of the injected false data. Furthermore, when the original received data $\boldsymbol{y}_0$ only contains a small amount of corruptions, $\boldsymbol{x}_0$ synthesized by the diffusion model can be regarded as the recovery results. Collectively, according to Eq. \eqref{gradient}, the corrected noise term can be rewritten as 
\begin{equation}
	\begin{aligned}
		\hat{\epsilon } = \epsilon_{\theta}(\boldsymbol{x}_{\tau_i},\tau_i)-\omega \sqrt{1-\alpha_{\tau_i}}(\boldsymbol{y}_{\tau_i}-\boldsymbol{x}_{\tau_i}),
	\end{aligned}
\end{equation}
where $\omega$ is the classifier guidance scaling parameter. When $\omega > 1$, the effect of conditional control is strong, and the measurements $\boldsymbol{x}_{\tau_{i-1}}$ will be quickly generated according to the expected trajectory. Nonetheless, if $\omega$ is too large, it is easy to make the synthesized reconstructed $\boldsymbol{x}_0$ the same as the input $\boldsymbol{y}_0$, which cannot achieve the purpose of anomaly detection and measurement recovery. To this end, the conditioned diffusion model should choose an appropriate $\omega$ for different scale power systems by conducting recovery error tests under different $\omega$.

\subsection{Diffusion-Based Imputation} \label{diffusionimputation}
Incomplete $\bm{y}_0$ can also guide the generation of diffusion models by exploiting spatial-temporal correlations to complement the missing positions, as discussed in subsection \ref{DL}. Denote the index set of correctly observed data points as $\Omega$, and the index set of abnormal or unobserved data points as $1-\Omega$. In this way, the known measurements can be represented as $\Omega \odot \boldsymbol{x}$, and the unidentified data can be described as $(1-\Omega) \odot \boldsymbol{x}$. Evidently, in every iteration of the reconstruction process, the normal measurements should remain unchanged, while the missing entries will be imputed by the generated data. The distribution of imputation part of data and the combined data can be expressed as 
\begin{equation}
	\begin{aligned}
			&\boldsymbol{x}^{\Omega}_{\tau_{i-1}} \sim \mathcal{N} (\sqrt{\alpha_{\tau_{i-1}}}\boldsymbol{x}_0,(1-\alpha_{\tau_{i-1}})\boldsymbol{I})\\
			&\boldsymbol{x}^{(1-\Omega)}_{\tau_{i-1}}\sim \mathcal{N}(\bar{\mu}(\boldsymbol{x}_{\tau_i}), \gamma^2_{\tau_i} \bar{\sigma}^2_{\tau_i}\boldsymbol{I})\\
		   &\boldsymbol{x}_{\tau_{i-1}}=\Omega\odot \boldsymbol{x}^{\Omega}_{\tau_{i-1}}+(1-\Omega)\odot \boldsymbol{x}^{(1-\Omega)}_{\tau_{i-1}} .
	\end{aligned}
\end{equation}

Additionally, in order to achieve better coordination between generated measurements and the normal observations, resampling is an appropriate enhancement method to obtain more coordinated results. In detail, $\boldsymbol{x}_{\tau_i}$ is obtained again by adding noises based on the denoising data $\boldsymbol{x}_{\tau_{i-1}}$, which can be represented as 
\begin{equation}
	\begin{aligned}
		\boldsymbol{x}_{\tau_i}=\sqrt{1-\beta_{\tau_{i}} }\boldsymbol{x}_{\tau_{i-1}}+\sqrt{\beta_{\tau_{i}}}\epsilon,
	\end{aligned}
\end{equation}
where $\epsilon \sim \mathcal{N} (\bm{0},\boldsymbol{I})$. Subsequently, the reverse process is performed on $\boldsymbol{x}_{\tau_i}$ again to compute $\boldsymbol{x}_{\tau_{i-1}}$. The proposed improved diffusion model will repeat the above resampling process by $R$ times to pursue higher accuracy and reliability.

\begin{figure}[!h]
	\vspace{-0.2cm}
	\centerline{\includegraphics[width=0.49\textwidth]{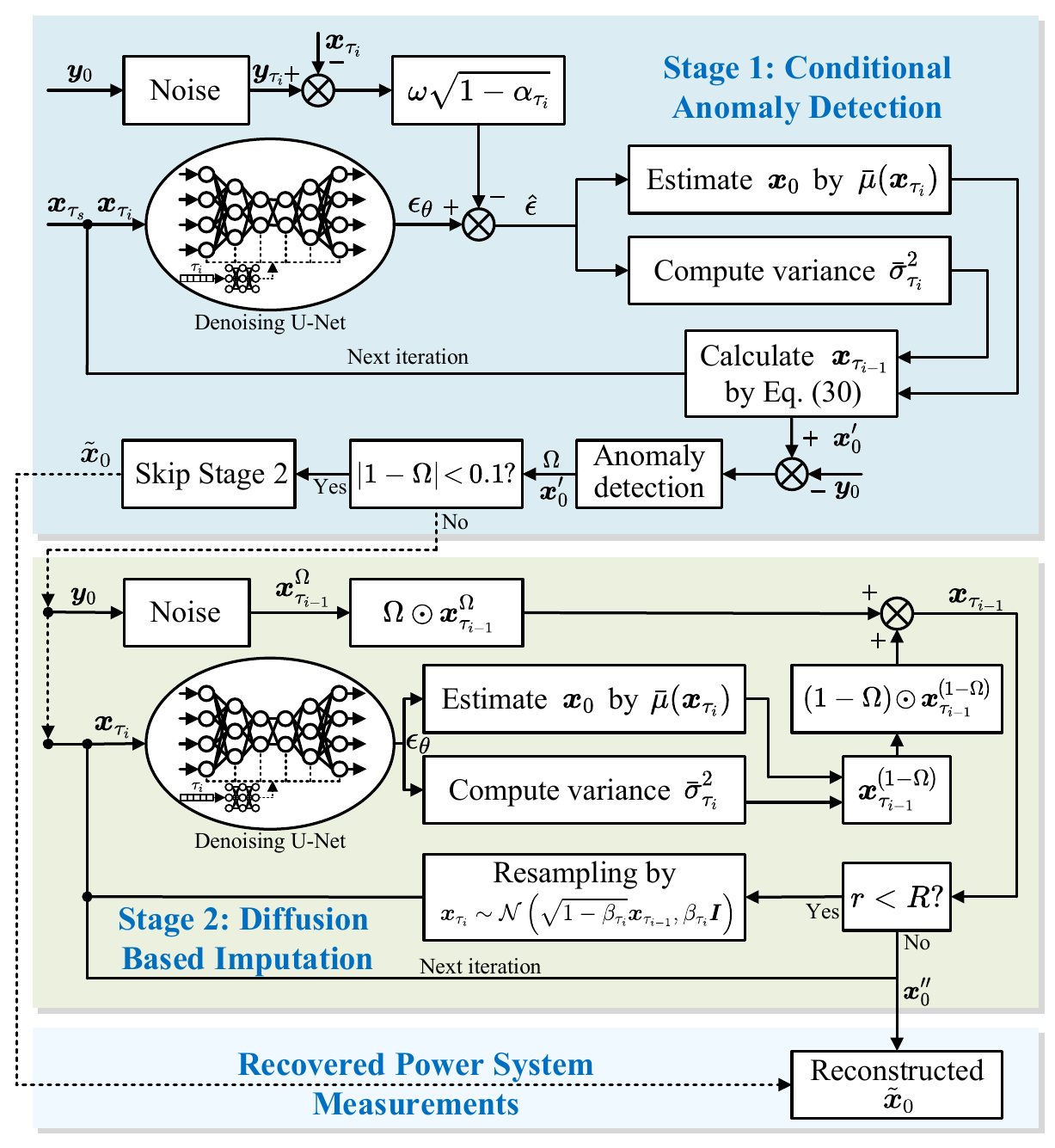}}
	\caption{The flowchart of improved efficient TSDM with Stage 1: conditional anomaly detection and Stage 2: diffusion based imputation.}
	\label{framework}
	\vspace{-0.2cm}
\end{figure}

\subsection{Improved Two-Stage Diffusion Recovery Model} \label{ITSDM}

The unreasonable setting of $\omega$ in different scenarios and the possible existence of massive false data and data losses will have a negative impact on restored data of the diffusion model, and $\boldsymbol{y}_{\tau_i}$ cannot effectively guide the precise data generation process. To this end, as illustrated in Fig. \ref{framework} and Algorithm \ref{Two-StageDiffusion}, the improved TSDM proposed in this paper consists of two stages: Stage 1 comprises a classifier-guided conditional anomaly detection model, which first uses the conditional diffusion technique discussed in subsection \ref{conditionaldiffusion} to recover a data matrix ${\bm{x}}'_0 \in \mathbb{R}^{M \times T}$ and then computes the difference between ${\bm{x}}'_0$ and the corrupted power system measurement matrix $\bm{y}_0$. Suspected outliers in $\bm{y}_0$ are then identified by performing 3$\sigma$ tests \cite{outlieranalysis} on the standard deviations. If the proportion of suspected outliers is less than 10\%, $\bm{x}'_0$ can be directly regarded as the output data $\tilde{\bm{x}}_0$. Otherwise, the index set $1 - \Omega$ of the suspected outliers is sent to Stage 2 with the same denoising U-Net, where the diffusion-based imputation discussed in subsection \ref{diffusionimputation} is used to restore the missing data and finally produce the output $\tilde{\bm{x}}_0$. In addition, the two-stage diffusion model utilizes DDIM and optimal variance acceleration strategy to achieve efficient data recovery.

\begin{algorithm}  	\label{Two-StageDiffusion}
	\footnotesize
	\caption{Improved efficient two-stage denoising diffusion measurement recovery model}
		\LinesNumbered
		\KwIn{Measurements matrix $\boldsymbol{y}_0 \in \mathbb{R}^{M \times T}$, variance shcedule $\{ \beta_n \} ^N_{n=1}$ and $\{ \alpha_n \}^N_{n=1}$, subsequence $\boldsymbol{\tau}=[\tau_1,\tau_2, \dots,\tau_s]$, guidance scaling parameter $\omega$, resampling parameter $R$}
		\KwOut{Restored measurements matrix $\tilde{\boldsymbol{x}}_0$}
		\textbf{Initialize} $\boldsymbol{x}$ by $\boldsymbol{x}_{\tau_s} \sim \mathcal{N}(\bm{0},\boldsymbol{I})$\;
		\textbf{Stage 1:}\\
		\For{all $i=s,\dots,2$}{
            \textbf{Compute} noisy data $\boldsymbol{y}_{\tau_i} \leftarrow \sqrt{\alpha_{\tau_i}}\boldsymbol{y}_0+ \sqrt{1-\alpha_{\tau_i}}\epsilon_{\theta}(\boldsymbol{x}_{\tau_i},\tau_i)$\;
			\textbf{Rectify} noise term $\hat{\epsilon} \leftarrow \epsilon_{\theta}(\boldsymbol{x}_{\tau_i},\tau_i)-\omega\sqrt{1-\alpha_{\tau_i}}(\boldsymbol{y}_{\tau_i}-\boldsymbol{x}_{\tau_i})$\;
			\textbf{Estimate} $\boldsymbol{x}_0$ by $\bar{\mu}(\boldsymbol{x}_{\tau_i}) \leftarrow \frac{1}{\sqrt{\alpha_{\tau_i}}}[\boldsymbol{x}_{\tau_i}-\sqrt{1-\alpha_{\tau_i}} \hat{\epsilon}]$\;
			\textbf{Compute} optimal variance by $\bar{\sigma}^2_{\tau_i} \leftarrow \frac{1-\alpha_{\tau_i}}{\alpha_{\tau_i}}\left \{ 1-\frac{1}{d} \mathbb{E}_{\boldsymbol{x}_{\tau_i}\sim p(\boldsymbol{x}_{\tau_i})}[\left \| \hat{\epsilon } \right \|^2_2] \right \}$\;
			\textbf{Update} $\boldsymbol{x}_{\tau_{i-1}}$ by Eq. \eqref{updatedata} using $\bar{\mu}(\boldsymbol{x}_{\tau_i})$ and $\bar{\sigma}^2_{\tau_i} $\;
		}
        \textbf{Estimate} $\boldsymbol{x}_0$ by ${\boldsymbol{x}}'_0 \leftarrow \frac{1}{\sqrt{\alpha_{\tau_1}}} [\boldsymbol{x}_{\tau_1}-\sqrt{1-\alpha_{\tau_1}}\hat{\epsilon}]$\;
		\textbf{Find} the outlier positions by $(1-\Omega) \leftarrow (\left | {\boldsymbol{x}}'_0 - \boldsymbol{y}_0 \right |>3\sigma_{\boldsymbol{y}_0} )$\;
		\If{ the proportion of outliers meets $\left | 1-\Omega \right | <0.1$}{
			\textbf{Output} $\tilde{\boldsymbol{x}}_0 \leftarrow {\boldsymbol{x}}'_0$ and \textbf{Skip} Stage 2\;
		}
		\textbf{Stage 2:}\\
		\For{all $i=s,\dots, 2$}{
			\For{$r=1, \dots, R$}{
			\textbf{Compute} known $\boldsymbol{x}^{\Omega}_{\tau_{i-1}} \leftarrow \sqrt{\alpha_{\tau_{i-1}}}\boldsymbol{y}_0 + \sqrt{1-\alpha_{\tau_{i-1}}}\epsilon_1$\;	
			\textbf{Estimate} $\boldsymbol{x}_0$ by $\bar{\mu}(\boldsymbol{x}_{\tau_i}) \leftarrow \frac{1}{\sqrt{\alpha_{\tau_i}}}[\boldsymbol{x}_{\tau_i}-\sqrt{1-\alpha_{\tau_i}} \epsilon_{\theta}(\boldsymbol{x}_{\tau_i},\tau_i)]$\;
			\textbf{Compute} optimal variance by $\bar{\sigma}^2_{\tau_i} \leftarrow \frac{1-\alpha_{\tau_i}}{\alpha_{\tau_i}} \left \{ 1-\frac{1}{d}\mathbb{E}_{\boldsymbol{x}_{\tau_i}\sim p(\boldsymbol{x}_{\tau_i})}[\left \| \epsilon_{\theta}(\boldsymbol{x}_{\tau_i},\tau_i) \right \|^2_2] \right \}$\;
			\textbf{Update} $\boldsymbol{x}_{\tau_{i-1}}^{(1-\Omega)}$ by Eq. \eqref{updatedata2} using $\bar{\mu}(\boldsymbol{x}_{\tau_i})$ and $\bar{\sigma}^2_{\tau_i} $\;
			\textbf{Combine} by $\boldsymbol{x}_{\tau_{i-1}} \leftarrow \Omega\odot \boldsymbol{x}^{\Omega}_{\tau_{i-1}}+(1-\Omega)\odot \boldsymbol{x}^{(1-\Omega)}_{\tau_{i-1}} $\;
			\If{$r<R$}{
				\textbf{Resample} $\boldsymbol{x}_{\tau_i} \leftarrow \sqrt{1-\beta_{\tau_{i}} }\boldsymbol{x}_{\tau_{i-1}}+\sqrt{\beta_{\tau_{i}}}\epsilon_2$\;
			}
			}
		}
        \textbf{Estimate} $\boldsymbol{x}_0$ by ${\boldsymbol{x}}'' _0 \leftarrow \Omega\odot \sqrt{\alpha_{\tau_1}}\boldsymbol{y}_0+(1-\Omega)\odot \bar{\mu}(\boldsymbol{x}_{\tau_1})$\;
	    \textbf{Return} reconstructed $\tilde{\boldsymbol{x}}_0 \leftarrow {\boldsymbol{x}}'_0/{\boldsymbol{x}}''_0 $

\end{algorithm}

Ignoring other simple computations, the time complexity of each noise $\epsilon_{\theta}$
prediction by denoising U-Net is $ \mathcal{O} \left ( \sum^{D}_{l=1} m^2_l t^2_l \cdot C_l \cdot C_{l-1} \cdot K^2_l \right ) $, where $D$ is the number of layers, $l$ represents the $l$-th layer, $m_ln_l$ denotes the feature size, $C_l$ and $C_{l-1}$ are the number of convolutional kernels in the $l$-th and $l-1$-th layer, and $K$ is the edge length of the convolutional kernel. The proposed TSDM needs to execute noise prediction $(s + R \cdot s)$ times in data recovery. The computational complexity of TSDM is $(1+R) \cdot s \cdot \mathcal{O} \left ( \sum^{D}_{l=1} m^2_l t^2_l \cdot C_l \cdot C_{l-1} \cdot K^2_l \right )$. Therefore, when the parameters and structure of the network are fixed, TSDM can reduce the computational complexity by minimizing $\textrm{NFE}$, selecting appropriate and sufficiently small $R$, and limiting the input size of $M\times T$.

\subsection{Cyber-Physical Implementation of TSDM} \label{CPITSDM}
\begin{figure}[!h]
	\vspace{-0.2cm}
	\centerline{\includegraphics[width=0.49\textwidth]{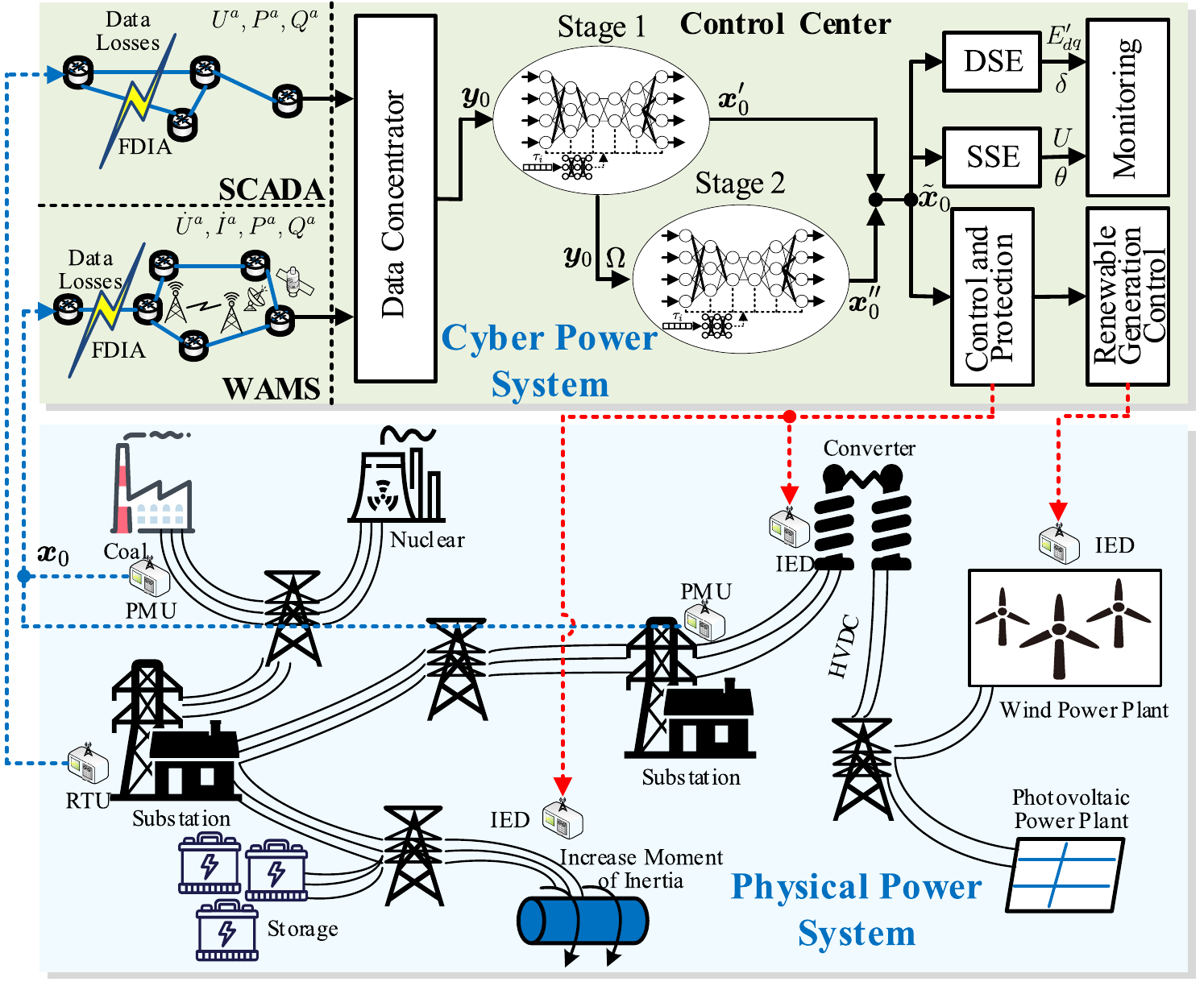}}
	\caption{Procedures and applications of the proposed TSDM algorithm.}
	\label{application}
	\vspace{-0.2cm}
\end{figure}

As illustrated in Fig. \ref{application}, on one hand, modern power systems estimate the partial static state $\boldsymbol{s}_k^{\mho}$ at time $k$ in the control center, and on the other hand, monitor the dynamic transients through WAMS. Measurements $\boldsymbol{z}_k^{\mho}$ collected by PMUs and RTUs during data transmission may be injected with false data $\Delta (\boldsymbol{z}_k^a)^{\mho }$ by hackers, and data losses may also occur due to communication malfunction and latencies, resulting in a corrupted or unobserved estimated state $(\boldsymbol{s}_k^a)^{\mho }$. The proposed TSDM can rectify biased static state variables through $\tilde{\boldsymbol{s}}_k^{\mho} = \mathop{\arg\min}_{\tilde{\boldsymbol{s}}_k^{\mho}} \left \| f^s_{\textrm{TSDM}}\left ( (\boldsymbol{z}_k^a)^{\mho} \right ) - h^{\mho}\left ( \tilde{\boldsymbol{s}}_k^{\mho} \right ) \right \|_0  $ by considering measurement equation constraints in power system area $\mho$, where $f^s_{\mathrm{TSDM}}(.)$ is the nonlinear function of TSDM with a $s$-length subsequence. Similarly, the correction step of various Kalman filtering methods can also accurately track the dynamic states with recovered measurements. Alternatively, the prior model can also be directly and flexibly established based on known state variables.

SSE/DSE in the control center can estimate the states $\boldsymbol{s}_{k,i}^{\mho} =  \left [ \theta_i,U_i \right ]^\top$ of the $i$-th Bus in SCADA or $\boldsymbol{s}_{k,i}^{\mho} =  \left [ \delta_{Gi}, \omega_{Gi}, E'_{dGi},E'_{qGi} , \theta_i, U_i \right ]^\top$ of the $i$-th Generator and its terminal bus in WAMS based on the input measurements. The proposed TSDM can eliminate the impact of contaminated data $\dot{\bm{U}}^a, \dot{\bm{I}}^a, \bm{P}^a, \bm{Q}^a$ and obtain accurate system static or dynamic states $\tilde{\boldsymbol{s}}_k^{\mho}$. Eventually, the rectified $\tilde{\boldsymbol{x}}_0$ and state variables $\tilde{\boldsymbol{s}}_k^{\mho}$ are utilized by modern power systems as the data support for monitoring, protection, and control, facilitating the safety and stability of complex, highly nonlinear power systems with measurement uncertainties.

% section Improved (end)

\section{Case Studies}  %(fold)
\label{sec:CS}
\subsection{Experimental Setup}

\begin{figure}[!h]
	\vspace{-0.2cm}
	\centerline{\includegraphics[width=0.49\textwidth]{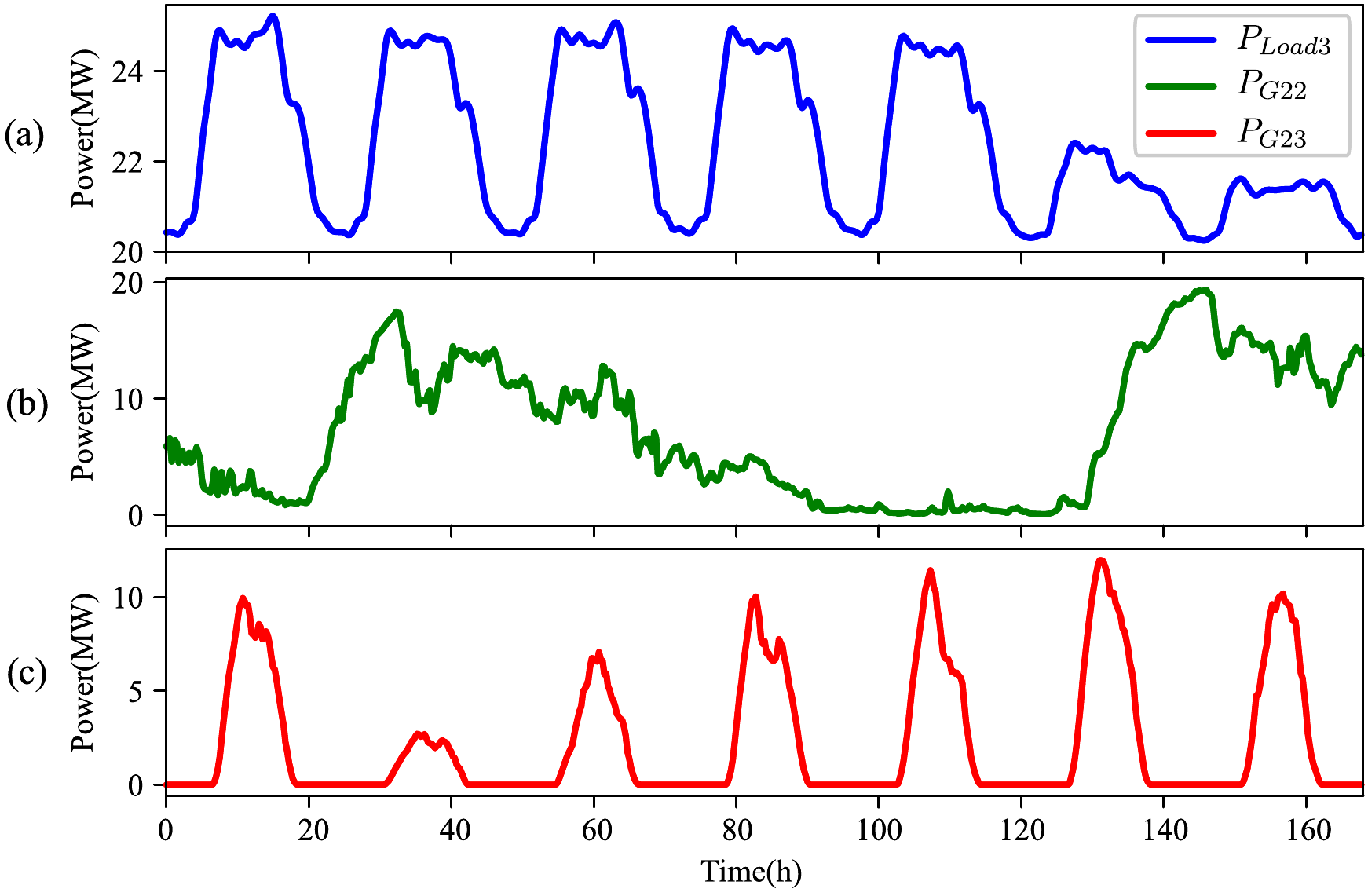}}
	\caption{The weekly commercial active power demand $\bm{P}_{Load3}$ of load 3 (a), power output $\bm{P}_{G22}$ of wind turbines 22 (b), and power output $\bm{P}_{G23}$ of photovoltaic generators 23 (c) in the IEEE 30-bus test system.}
	\label{dataset}
	\vspace{-0.2cm}
\end{figure}

\textit{1) Dataset Source:} Two typical application scenarios, i.e., the RTU-based SCADA system and the PMU-based WAMS, are implemented for case studies. For the SCADA scenario, training and test data of IEEE 30, 57, and 118-bus systems are obtained via simulation on MATPOWER 7.0. The load changes $\Delta \bm{P}_L$ are extracted from \cite{9174809}. Furthermore, parts of generation profiles $\Delta \bm{P}_G$ are replaced with actual renewable power output from January 2022 to June 2023 in Belgium \cite{9305967}. The load and renewables output curve $[\bm{P}_L^{\top}, \bm{P}^{\top}_G]^{\top}$ for a certain week is illustrated in Fig. \ref{dataset}. The data reporting rate is set to 1 sample per 15 minutes. For the WAMS scenario, transient data of IEEE 39-bus and NPCC 140-bus systems are obtained using the open-source simulator Python-based ANDES \cite{9169830}. PMUs are installed at 13 optimal locations in IEEE 39-bus system \cite{4519389} and 30 generator terminal buses in NPCC 140-bus system \cite{6913022}, with a reporting rate of 30 Hz. Five thousand different power system events, including short-circuit faults, line trips, generator sheddings, and load changes, are considered to simulate the system input change $\Delta \bm{u}$.

\textit{2) Cyber-Physical Co-Simulation:} Measurement recovery experiments are performed based on the cyber-physical power system co-simulation platform with a Coordinator-Event-Axis-Based Time
Synchronization (CoordTS) Strategy \cite{10376222}. CoordTS supports the integration of multiple software such as PSCAD/EMTDC, MATLAB/MATPOWER, and Python. The communication protocols such as IEC 61850 and IEEE C37.118 are implemented in the network simulator 3 (NS3). Moreover, CoordTS can simulate multiple types of cyber attacks such as address resolution protocol (ARP), distributed denial of service (DDoS), and man in the middle attacks. Based on this, various types of measurement uncertainty in Subsection \ref{DL} can be implemented in CoordTS.

\textit{3) Model Training:} An independent TSDM is trained for each test system based on Pytorch 1.8.1. The total timesteps and DDIM subsequence lengths are uniformly set to $N$=100 and $s$=10, respectively. Meanwhile, LSTM, variational autoencoder (VAE), and GAN are also trained for comparison, where the structure of the GAN follows the optimized Wasserstein GAN (WGAN) with an additional encoder. The simulations run in MATLAB R2021a and Python 3.8.6 on a computer with an i7-8700U 3.2 GHz CPU and 16 GB of RAM.

\textit{4) Performance Metrics:} Weighted root mean squared error (RMSE) is utilized to verify the state reconstruction accuracy 
\begin{equation}
	\begin{aligned}
		\mathrm{RMSE}&=  \left \| \boldsymbol{m}\left ( \boldsymbol{s}^{\mho}_{1:T}- \tilde{\boldsymbol{s}}^{\mho}_{1:T}\right ) \right \|_2 \\
           &= \sqrt{\frac{1}{S\times T}\sum_{i=1}^{S}m_i\sum_{j=1}^{T} \left ( \boldsymbol{s}^{\mho}_{ij} - \tilde{\boldsymbol{s}}^{\mho}_{ij} \right )^2  },
	\end{aligned}
\end{equation}
where $\boldsymbol{s}^{\mho}_{ij}$ and $\tilde{\boldsymbol{s}}^{\mho}_{ij}$ are the actual and restored state elements, and $m_i$ is the weight coefficient of the $i$-th state variable.

\subsection{Steady-State Recovery of SCADA Measurements}
\begin{figure}[!h]
	\vspace{-0.1cm}
	\centerline{\includegraphics[width=0.49\textwidth]{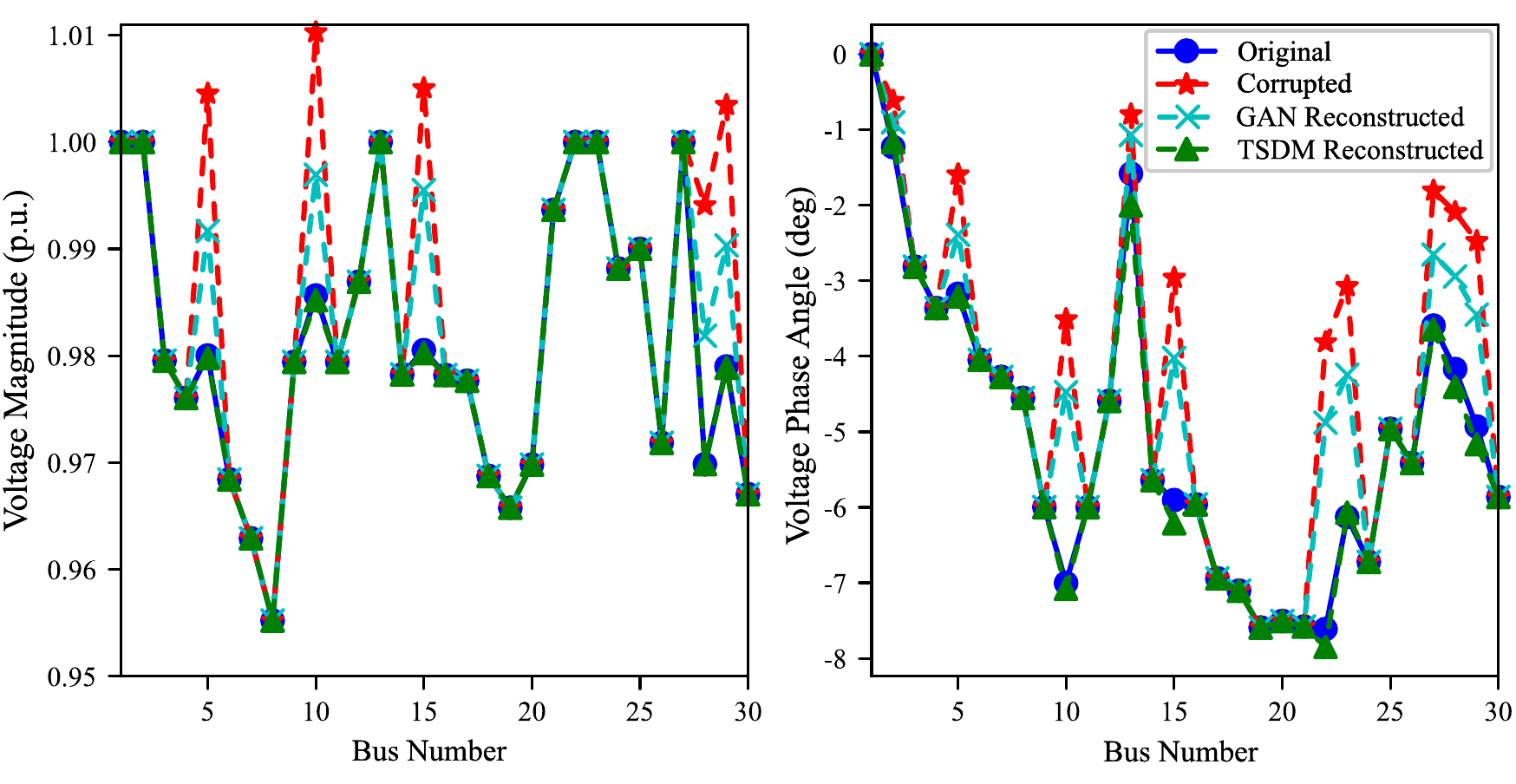}}
	\caption{The estimated state variables corrupted by step attacks and reconstructed by TSDM and GAN in IEEE 30-bus system.}
	\label{IEEE30}
	\vspace{-0.1cm}
\end{figure}

TSDM and GAN are first tested using the SCADA data from the IEEE 30-bus test system. The models are trained using the measurements of 2022, and are tested with the data from 6 am to 4 pm on Jan. 1, 2023. In the test data, step attacks are applied to 20\% of state variables, with the attack amplitude equal to 2.5\% for voltage magnitudes and 50\% for phase angles. The original and corrupted state estimates at 10 am, as well as the state estimates restored by TSDM and GAN, are shown in Fig. \ref{IEEE30}. It can be seen that the proposed TSDM achieves a higher data recovery accuracy, and the predicted output is generally consistent with the original state estimates. On the other hand, GAN still shows non-negligible deviations from the expected values, although the deviations have already been suppressed to some extent.

\begin{table*}[!t]\centering \footnotesize
	\caption{Weighted RMSE of IEEE 30-bus, 57-bus, and 118-bus in steady-state recovery via five methods \cite{8356714,8120129,8428530,9817514,8731755}}
	\label{SCADAresults}
	\renewcommand{\arraystretch}{1.0}
	\setlength\tabcolsep{0.5em}
	\begin{tabular}{ccccccccccccccccc}
	\toprule
	\toprule
	\multicolumn{2}{c}{Test System}     & \multicolumn{5}{c}{IEEE 30-bus} & \multicolumn{5}{c}{IEEE 57-bus} & \multicolumn{5}{c}{IEEE 118-bus} \\
	\cmidrule(lr){1-2} \cmidrule(lr){3-7}\cmidrule(lr){8-12} \cmidrule(lr){13-17}  
	\multicolumn{2}{c}{\diagbox{Anomalies}{Method}}         & LSTM  & ADMM & VAE & GAN & TSDM & LSTM  & ADMM & VAE & GAN & TSDM & LSTM  & ADMM  & VAE & GAN & TSDM \\
	\midrule
	\multirow{3}{*}{FDIA}      & Step   &      4.58 &    3.66  &  2.32   &   2.22  &   \textbf{0.77}   & 4.27     &  3.80    &  3.83   &   2.38  &  \textbf{0.78}    & 13.45      & 6.15      & 3.66    & 3.09    & \textbf{1.28}    \\
			     & Ramp   &   4.19    &  3.52    &    2.42 &  1.61   &  \textbf{0.55}    &  4.23     &  3.70    & 3.91    &   2.09  &  \textbf{0.72}    & 11.44      &  6.07     & 3.63    &  2.99   &   \textbf{1.26}   \\
		                     & Random & 3.53      &   2.61   &   2.31  &   1.98  &    \textbf{0.42}  &  4.27     & 2.07     &  3.72   &  2.67   &  \textbf{0.57}    &   13.45    &  6.32     & 3.48    &   2.93  &  \textbf{1.14}    \\
								   \midrule
	\multirow{2}{*}{Data Loss} & RM     &   4.35    &   0.56   &   3.48  &  3.11   &    \textbf{0.14}  &  3.49     &   0.39   &  3.99   &   2.86  &  \textbf{0.39}    &  3.71     & 0.47      &  3.63   &   3.12  &  \textbf{0.23}    \\
			        & NM     &    2.20   &  0.65    &  3.54   &  2.51   &  \textbf{0.30}    &  2.92     & 0.39     &  3.98   &  2.19   &  \textbf{0.15}    &   3.37    &  0.51    &  3.60   & 3.48    &  \textbf{0.14} \\
	\bottomrule
	\bottomrule
	\end{tabular}
	\vspace{-0.1cm}
\end{table*}

\begin{figure}[!h]
	\vspace{-0.1cm}
	\centerline{\includegraphics[width=0.49\textwidth]{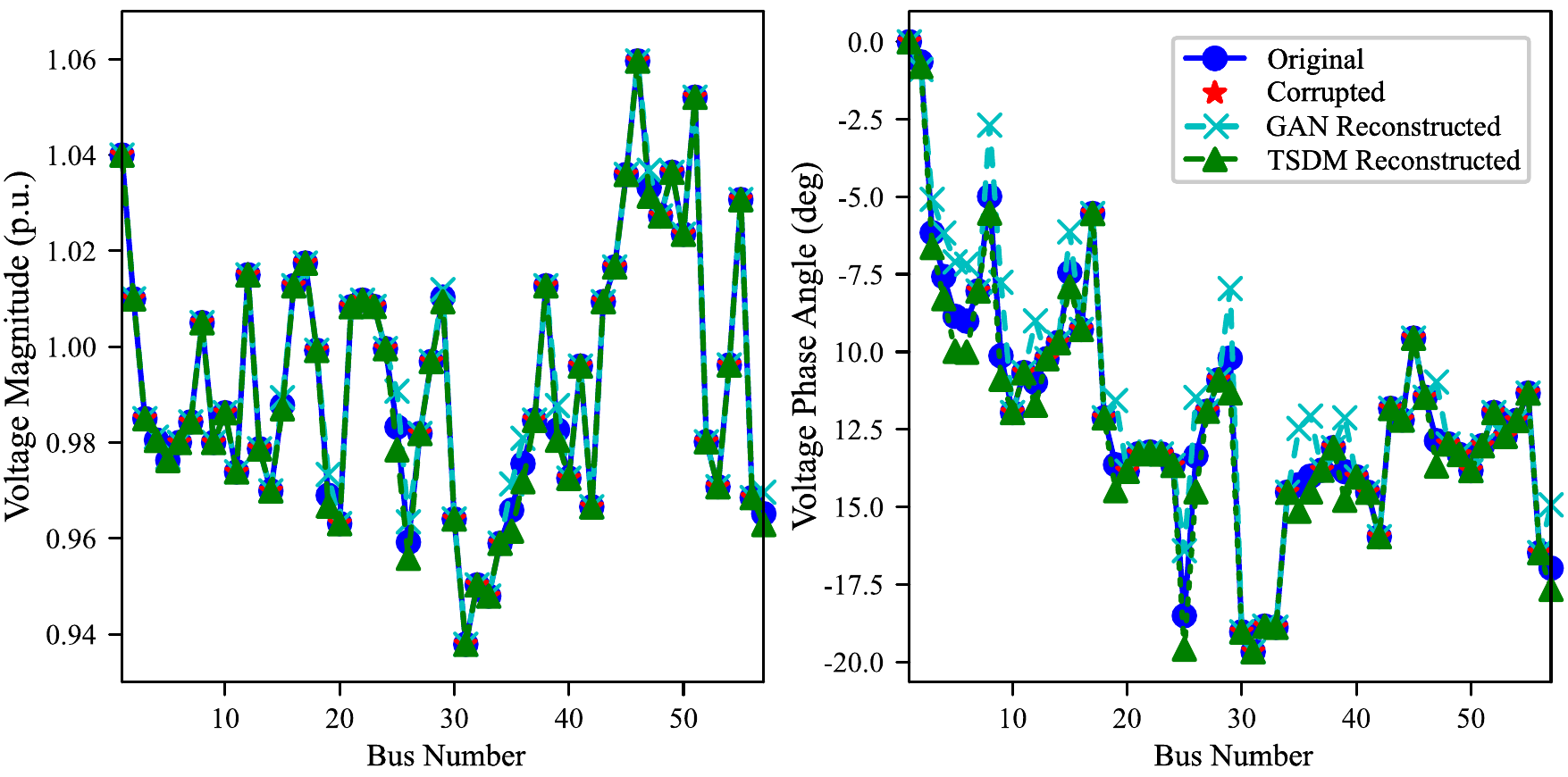}}
	\caption{The estimated state variables corrupted by NM and reconstructed by TSDM and GAN in IEEE 57-bus system.}
	\label{IEEE57}
	\vspace{-0.1cm}
\end{figure}

The second test assumes an NM data loss scenario between 8 am and 6 pm, where the measurements related to about 30\% unobservable states in the IEEE 57-bus system cannot reach the control center. The unobservable states are listed in Table \ref{tab:observability}.  The original and corrupted state estimates at 10 am, as well as the state estimates recovered by the two models, are shown in Fig. \ref{IEEE57}. Similar to the results above, TSDM apparently has smaller data imputation errors than GAN.

\begin{table}[!h]\centering \footnotesize
	\vspace{-0.1cm}
	\caption{The unobservable states of IEEE 57-bus system}
	\label{tab:observability}
	\renewcommand{\arraystretch}{1.0}
\renewcommand\tabcolsep{8pt}
	\begin{tabular}{cc}
	\toprule
	\toprule
	\multirow{3}{*}{\shortstack{Unobservable \\  States}} &   $U_4$, $U_5$, $U_{15}$, $U_{19}$, $U_{25}$, $U_{26}$, $U_{29}$, $U_{35}$, $U_{36}$, $U_{39}$, \\
	& $U_{47}$, $U_{57}$, $\theta_2$, $\theta_3$, $\theta_4$, $\theta_5$, $\theta_6$, $\theta_8$, $\theta_9$, $\theta_{12}$, $\theta_{15}$, \\
	&$\theta_{19}$, $\theta_{25}$, $\theta_{26}$, $\theta_{29}$, $\theta_{35}$, $\theta_{36}$, $\theta_{39}$, $\theta_{47}$,  $\theta_{57}$   \\
	\bottomrule
	\bottomrule
	\end{tabular}
	\vspace{-0.1cm}
	\end{table}

\begin{figure}[!h]
	\vspace{-0.1cm}
	\centerline{\includegraphics[width=0.49\textwidth]{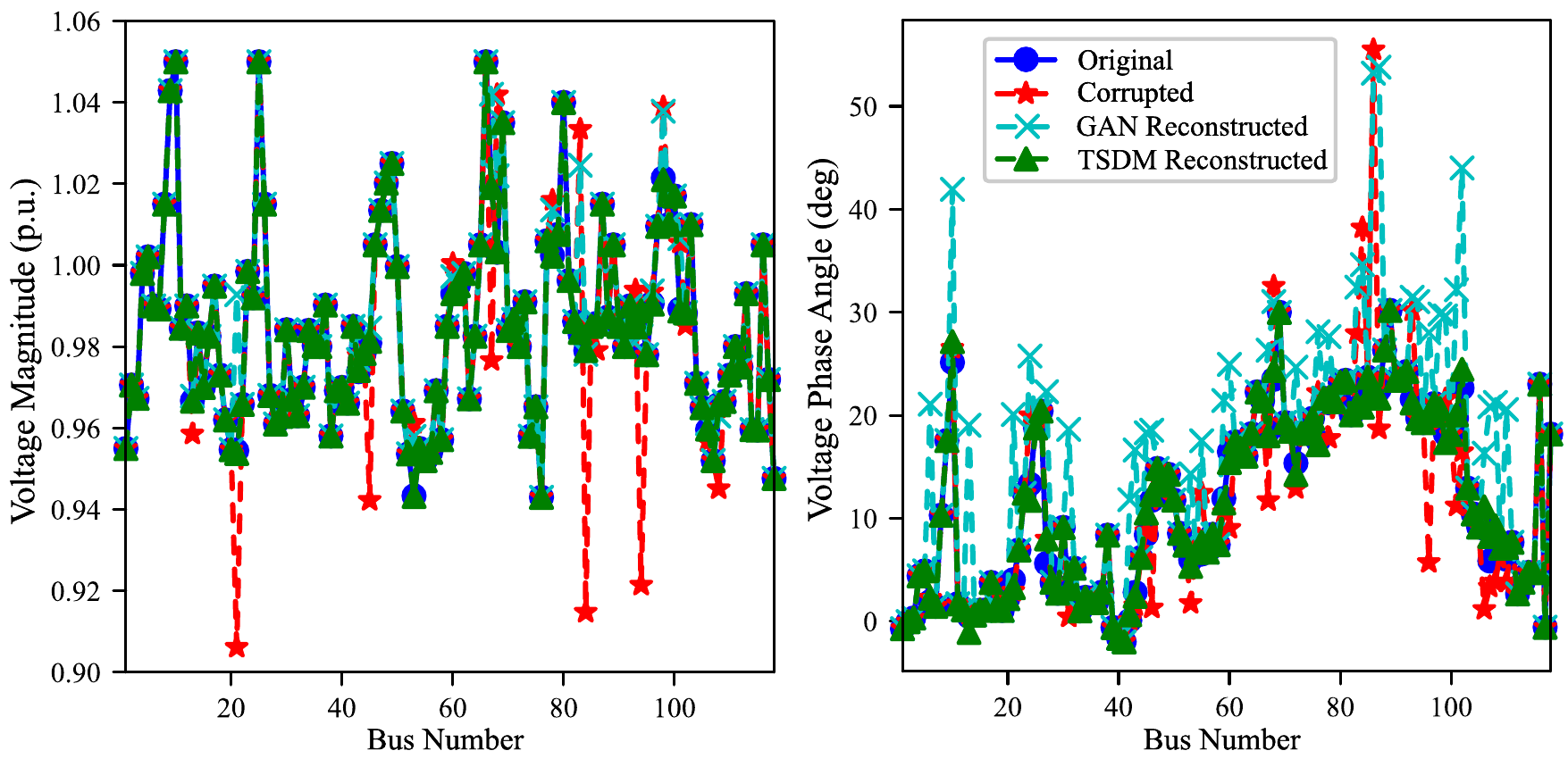}}
	\caption{The estimated state variables corrupted by random FDIA and reconstructed by TSDM and GAN in IEEE 118-bus system.}
	\label{IEEE118}
	\vspace{-0.1cm}
\end{figure}

For the large-scale 118-bus system, random FDIAs with zero mean, magnitude variance of 0.03, and phase angle variance of 0.6 are applied to 30\% of the state variables from 6 am to 8 pm. The original and corrupted state estimates and the estimates obtained based on restored data are shown in Fig. \ref{IEEE118}. Again, state variables can be accurately estimated using the data reconstructed by TSDM. However, GAN-restored data lead to a significant bias in phase angle estimation.

Numerical recovery experiments are carried out in different scenarios to verify the proposed TSDM. The test data include SCADA measurements sampled from January to June 2023 contaminated by step/ramp/random FDIA and RM/NM data loss with attack region modified ratios of 1\% to 50\%. The magnitude and phase angle attack amplitudes are 1\%-10\% and 10\%-100\%, respectively. The algorithms for comparison include the time-series prediction model LSTM \cite{8356714}, ADMM based robust principal component analysis and matrix completion mathematical algorithms \cite{8120129,8428530}, and reconstruction networks Gaussian mixture model based VAE \cite{9817514} and Encoder-WGAN \cite{8731755}. Table \ref{SCADAresults} shows that the time-series models represented by LSTM have difficulty in accurately predicting the quasi-steady measurement changes in modern power systems with renewable energy integration. Moreover, ADMM has a significant RMSE in FDIA data recovery tasks, and the RMSE of two typical reconstruction algorithms, VAE and GAN, is much greater than that of the proposed TSDM.

\subsection{Dynamic Recovery of PMU Measurements}
\begin{figure*}[!t]
	\centering   
	\subfigure[Data recovery results against a replay attack during a three phase fault.] 
	{
        \label{NPCC140replay}
		\includegraphics[width=0.47\linewidth]{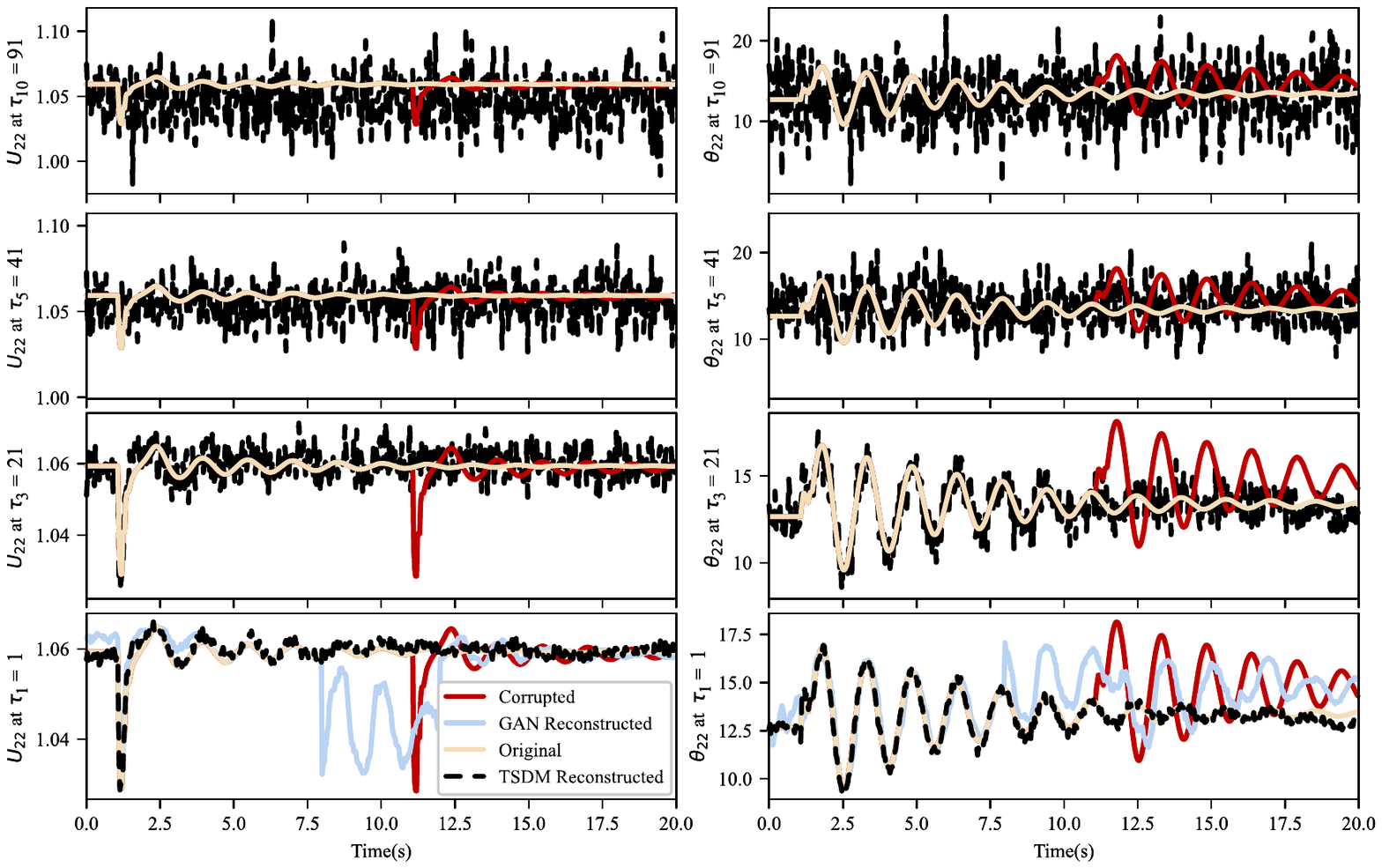}
	}
	\subfigure[Data recovery results against data loss during line trips.]
	{
        \label{IEEE39linetrip}
		\includegraphics[width=0.49\linewidth]{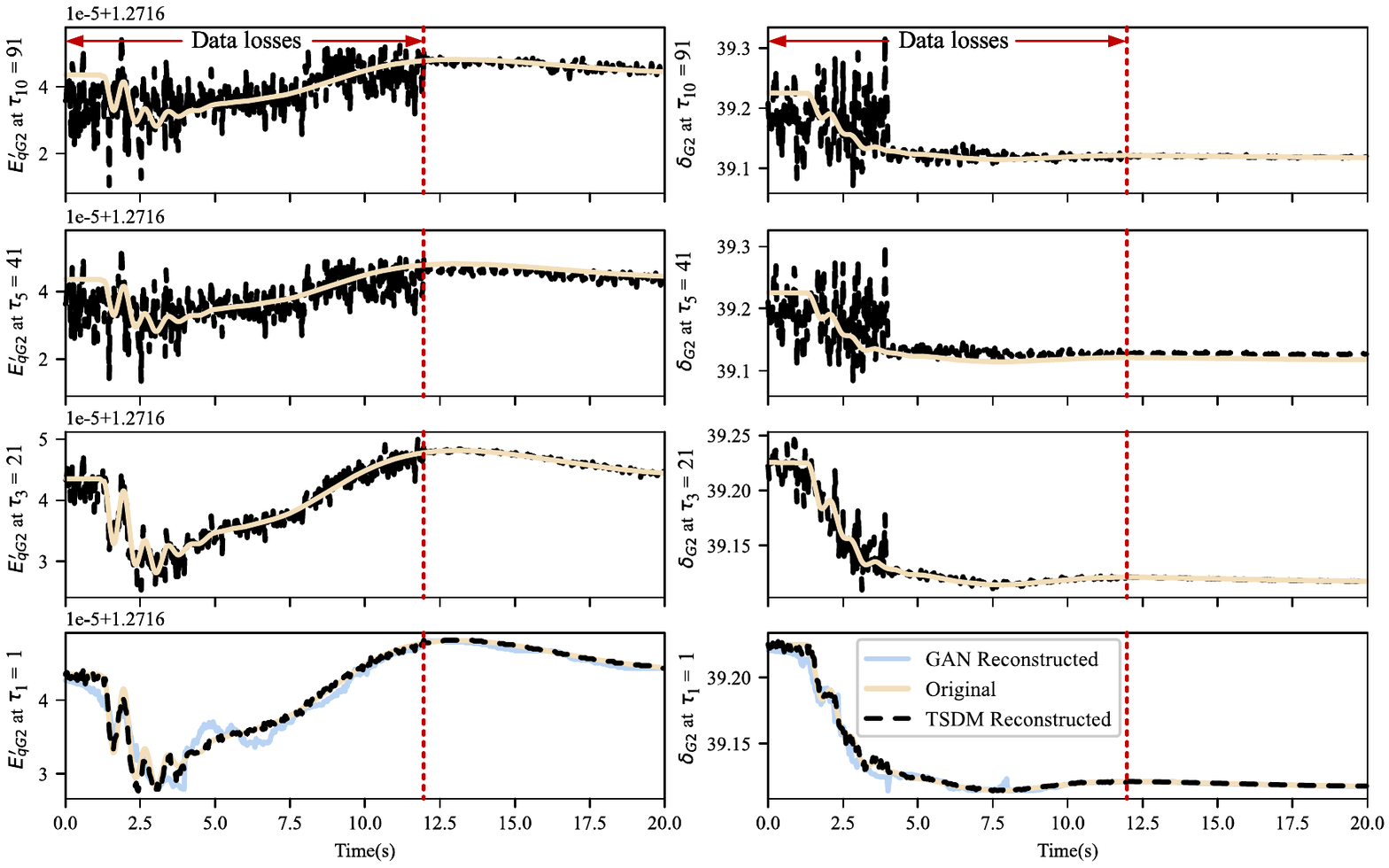}

	}
	\caption{The generated results of some selected steps of TSDM and GAN recovered states.}
\end{figure*}

To demonstrate the performance of TSDM in complex nonlinear dynamics, PMU data under a series of power system events are used in this set of tests. In the first case, a three-phase fault with grounding reactance $x_f$=0.075 p.u. occurs at 1s near Bus 5 of the NPCC 140-bus system and is cleared at 1.06s, and Line 5 between Buses 5 and 6 is disconnected at 1.05s and recloses at 1.104s. Moreover, a replay attack is applied during 11s-20s. The measurements of Bus 22 restored by GAN and TSDM are illustrated in Fig. \ref{NPCC140replay}. It can be seen that the output of TSDM gradually converges to the states along the iteration steps of 96, 46, 21, and 1 of the underlying DDPM model, while the GAN output is significantly affected by the contingencies.

In the second case, Line 11 between Buses 6 and 7 and Line 28 between Buses 22 and 23 in the IEEE 39-bus system are disconnected from 1s to 1.06s, with an NM data loss applied during 0s-12s. The measurements of Generator 2 during the line trips restored by GAN and TSDM are illustrated in Fig. \ref{IEEE39linetrip}. It can be seen that the missing entries of the power angle $\delta_{G2}$ and q-axis transient voltage $E'_{qG2}$ of Generator 2 gradually approach the actual values during the guided conditional denoising process of TSDM. However, the recovery accuracy of GAN is slightly worse.

\begin{table*}[!t]\centering \footnotesize
	\caption{Weighted RMSE of IEEE 39-bus and NPCC 140-bus in dynamic-state recovery via five methods \cite{8356714,8120129,8428530,9817514,8731755}}
	\label{WAMSresults}
	\renewcommand{\arraystretch}{1.0}
	\setlength\tabcolsep{1.2em}
	\begin{tabular}{cccccccccccc}
	\toprule
	\toprule
	\multicolumn{2}{c}{Test System}     & \multicolumn{5}{c}{IEEE 39-bus} & \multicolumn{5}{c}{NPCC 140-bus} \\
	\cmidrule(lr){1-2} \cmidrule(lr){3-7}\cmidrule(lr){8-12} 
	\multicolumn{2}{c}{\diagbox{Anomalies}{Method}}         & LSTM  & ADMM & VAE & GAN & TSDM & LSTM  & ADMM & VAE & GAN & TSDM  \\
	\midrule
	\multirow{3}{*}{FDIA}      & Step   &  3.07     &  2.67    &   1.01  & 0.42    &  \textbf{0.20}    & 3.67      & 3.41     &  1.40   &  0.68   &     \textbf{0.37}     \\
			     & Ramp   &  3.08     &  2.09    &  1.03   &  0.31   &   \textbf{0.19}   &   3.43    & 3.32    &  1.41   & 0.78    & \textbf{0.34}              \\
			     & Random &  2.97     & 3.04     &   1.01  &  0.74   &  \textbf{0.22}    & 3.12      & 2.90     &  1.38   &  0.55   &      \textbf{0.31}          \\
                \midrule
	\multirow{2}{*}{Data Loss} & RM     & 3.40      &  0.53    &   1.15  &  0.47   &  \textbf{0.14}    & 3.80      &  0.36   &  2.23   &  1.45   &  \textbf{0.21}            \\
			        & NM     &  3.25     &  0.55    &  1.87   & 1.10    &   \textbf{0.10}   & 3.88      &  0.31    &  2.15   & 1.40    &    \textbf{0.14}          \\
	\bottomrule
	\bottomrule
	\end{tabular}
	\vspace{-0.1cm}
	\end{table*}

\begin{figure*}[!t]
		\centering   
		\subfigure[A composite FDIA during a load change.] 
		{
			\label{IEEE39load}
			\includegraphics[width=0.48\linewidth]{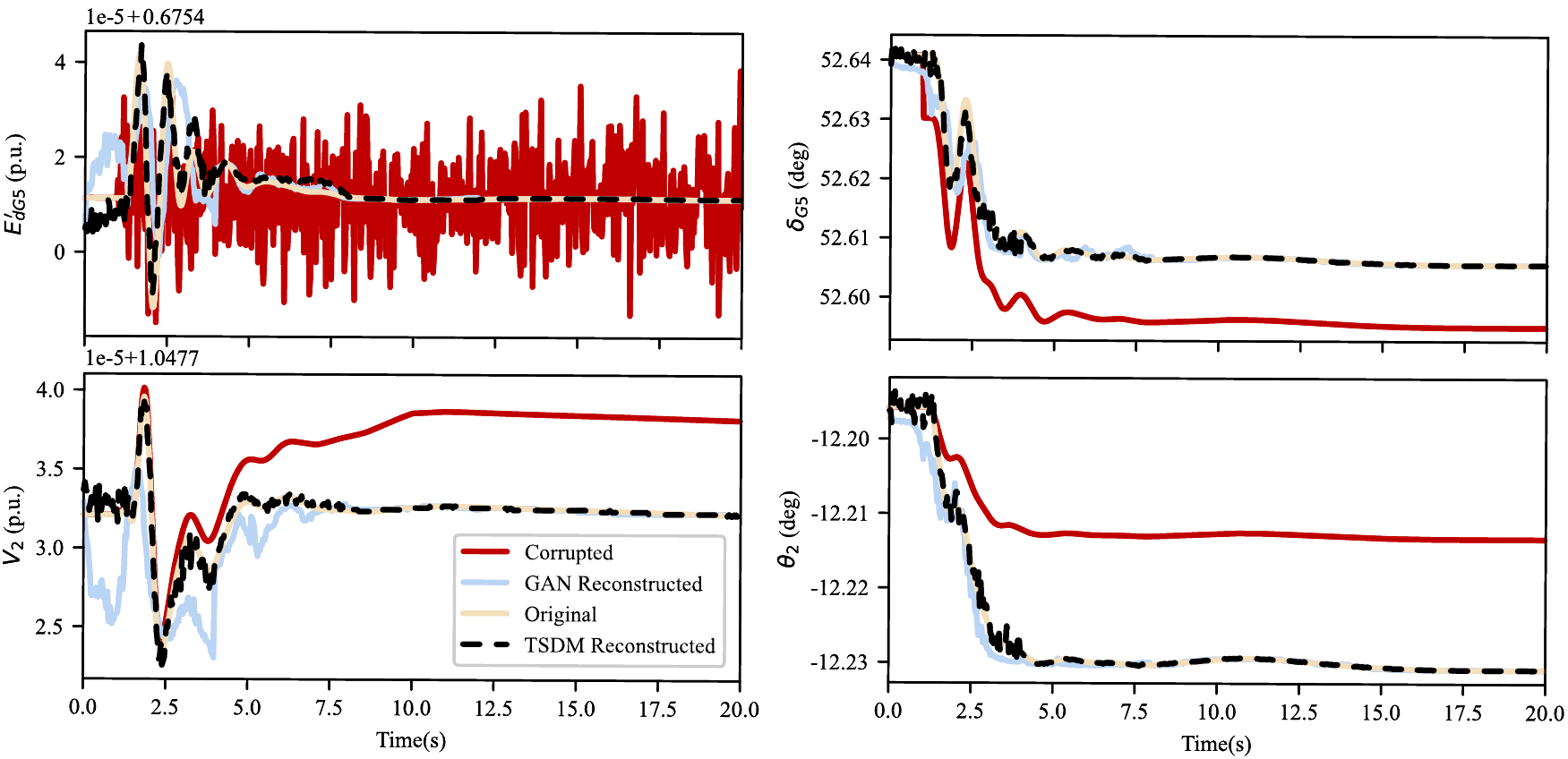}
		}
		\subfigure[Additive noises with varying variance during generator trips.]
		{
			\label{IEEE39varyingvariance}
			\includegraphics[width=0.48\linewidth]{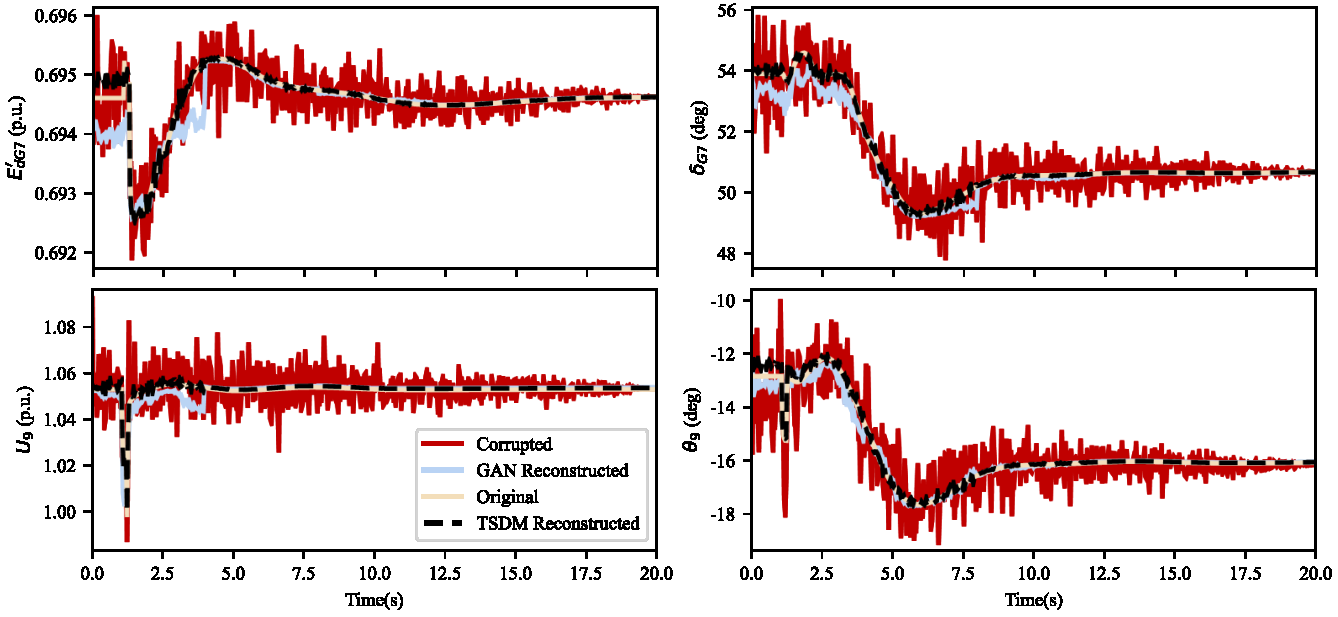}
	
		}
		\caption{TSDM and GAN recovered states in IEEE 39-bus system.}
\end{figure*}

\begin{figure*}[!t]
	\centering   
	\subfigure[Data losses during a generator shedding.] 
	{
		\label{NPCC140gentrip}
		\includegraphics[width=0.47\linewidth]{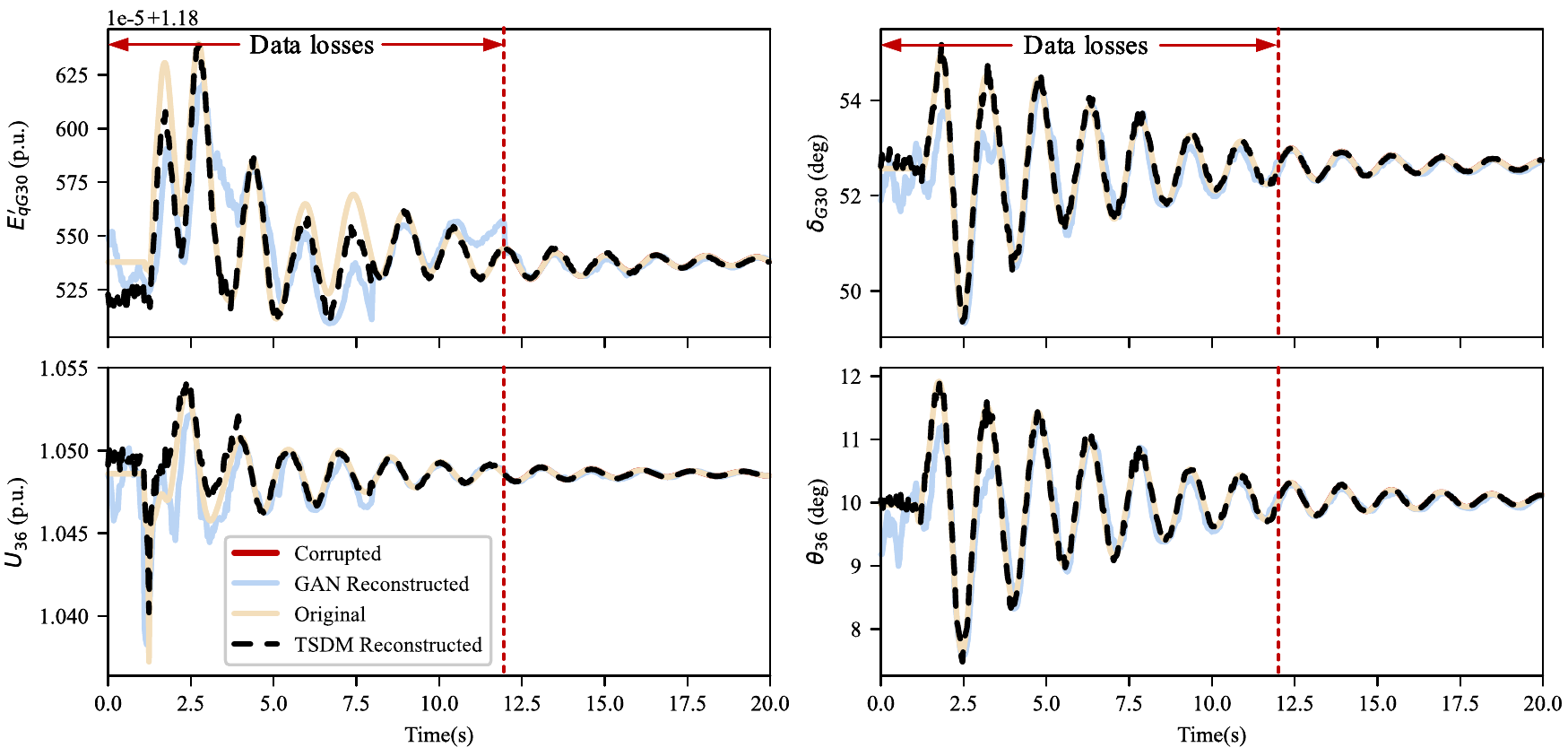}
	}
	\subfigure[$180^{\circ}$ phase shift attack during a line trip.]
	{
		\label{NPCC140phase}
		\includegraphics[width=0.49\linewidth]{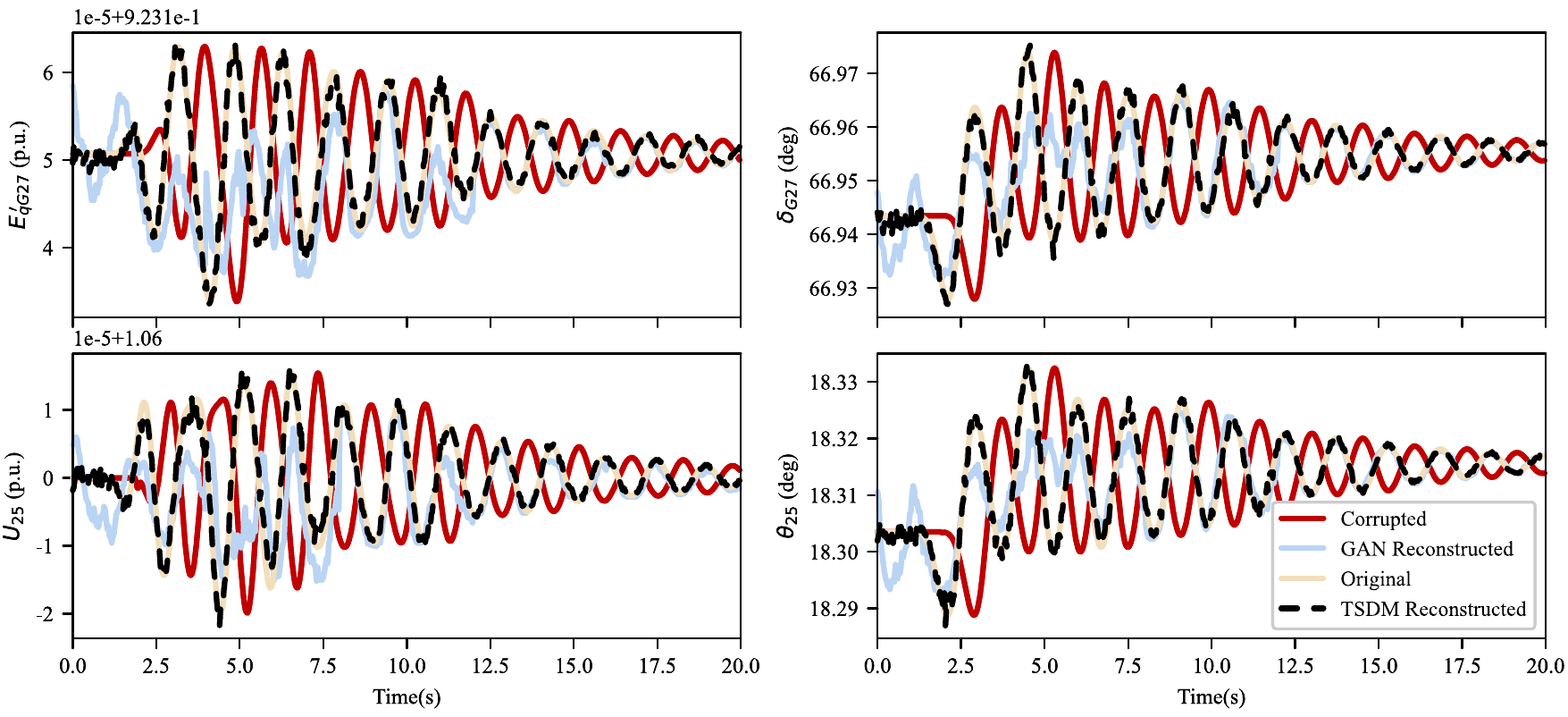}

	}
	\subfigure[A composite FDIA during a load change.]
	{
		\label{NPCC140load}
		\includegraphics[width=0.48\linewidth]{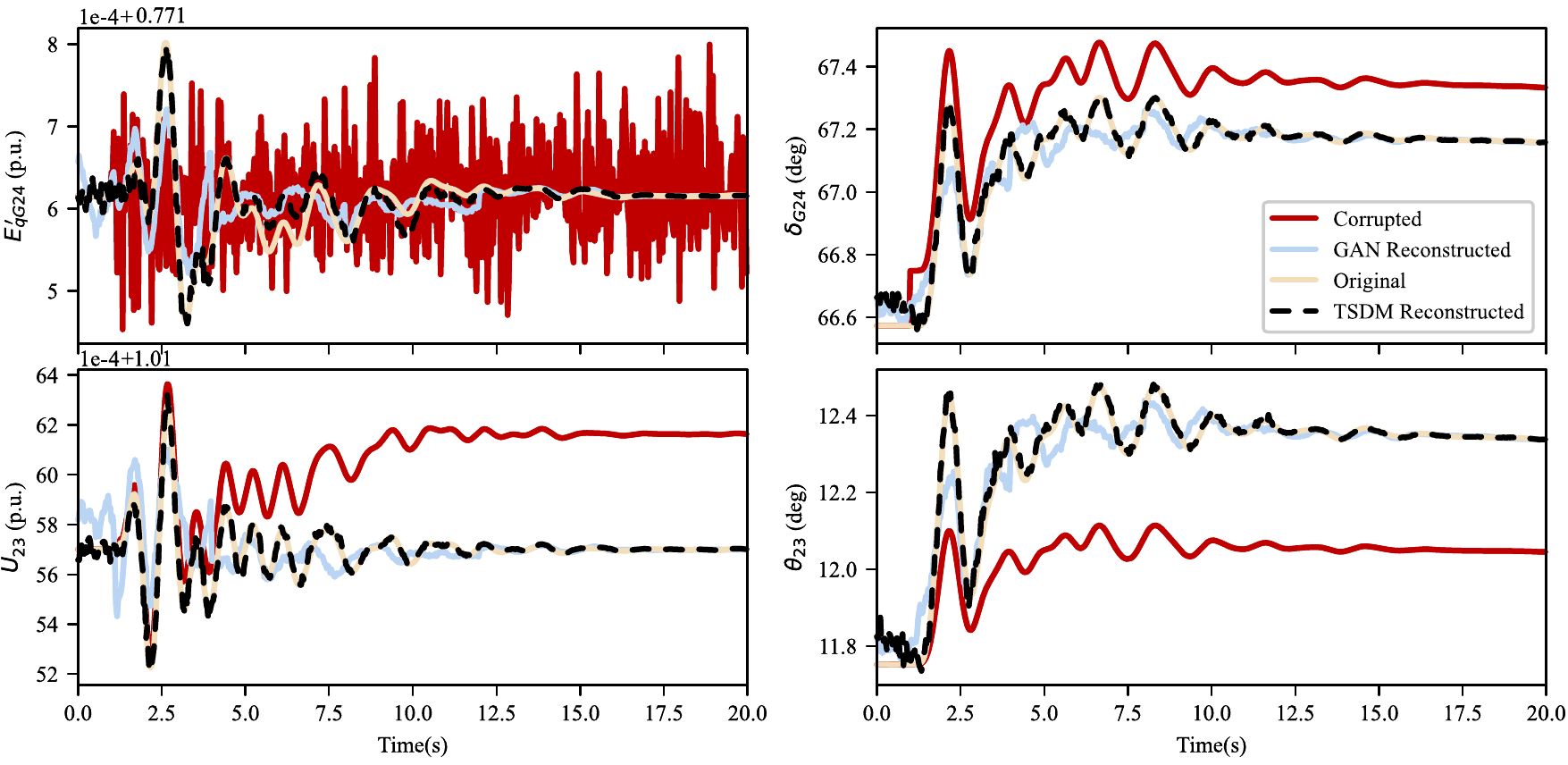}
	}
	\subfigure[Amplitude scaling and phase shift attack during a line trip with highly nonlinear dynamics.]
	{
		\label{NPCC140nonlinear}
		\includegraphics[width=0.48\linewidth]{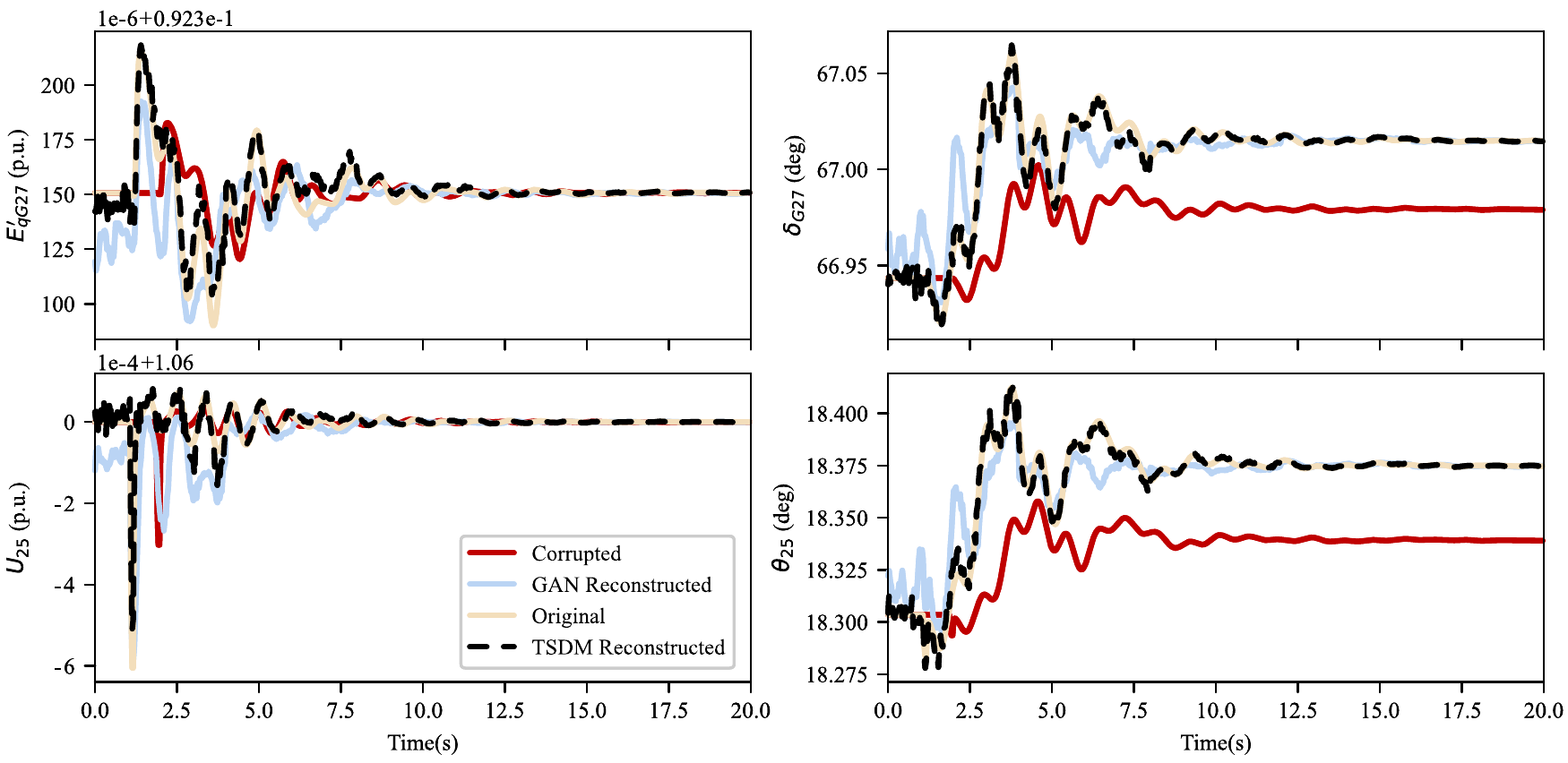}

	}
	\caption{TSDM and GAN recovered states in NPCC 140-bus system.}
\end{figure*}

The third case assumes a composite attack scenario to verify the recovery effectiveness of TSDM. In this case, a 68 MW load is disconnected to Bus 35 during 1s-1.05s, with random noises of zero mean and 0.3 variance, a 30\% step attack, an 80\% ramp attack, and a 50\% amplitude scaling attack applied to the transient d-axis voltage $E'_{dG5}$, power angle $\delta_{G5}$, bus voltage magnitude $V_2$, and phase angle $\theta_2$, respectively. Fig. \ref{IEEE39load} indicates that TSDM can still effectively restore the highly nonlinear dynamics in such a complex FDIA scenario. Moreover, TSDM can also accurately estimate state changes in the presence of additive noises with time-varying variance \cite{9721122} in Fig. \ref{IEEE39varyingvariance}. Similarly, TSDM can accurately recover measurements and ensure state estimation accuracy in generator trip events in Fig. \ref{NPCC140gentrip}, the scenario of phase shift attacks in Fig. \ref{NPCC140phase}, the scenario of a load change in Fig. \ref{NPCC140load}, and highly nonlinear dynamics in Fig. \ref{NPCC140nonlinear}.

The test data of the numerical verification are the WAMS measurements collected during different power system contingencies not included in the training datasets. The modified ratios are 1\% to 50\%. In Table \ref{WAMSresults}, the proposed TSDM still maintains and demonstrates strong superiority in power grid dynamics, with RMSE smaller than existing methods in both FDIA reconstruction tasks and unobservable state variables complementation.

\subsection{Hyperparameter Selection}
\begin{figure}[!h]
	\vspace{-0.1cm}
	\centerline{\includegraphics[width=0.49\textwidth]{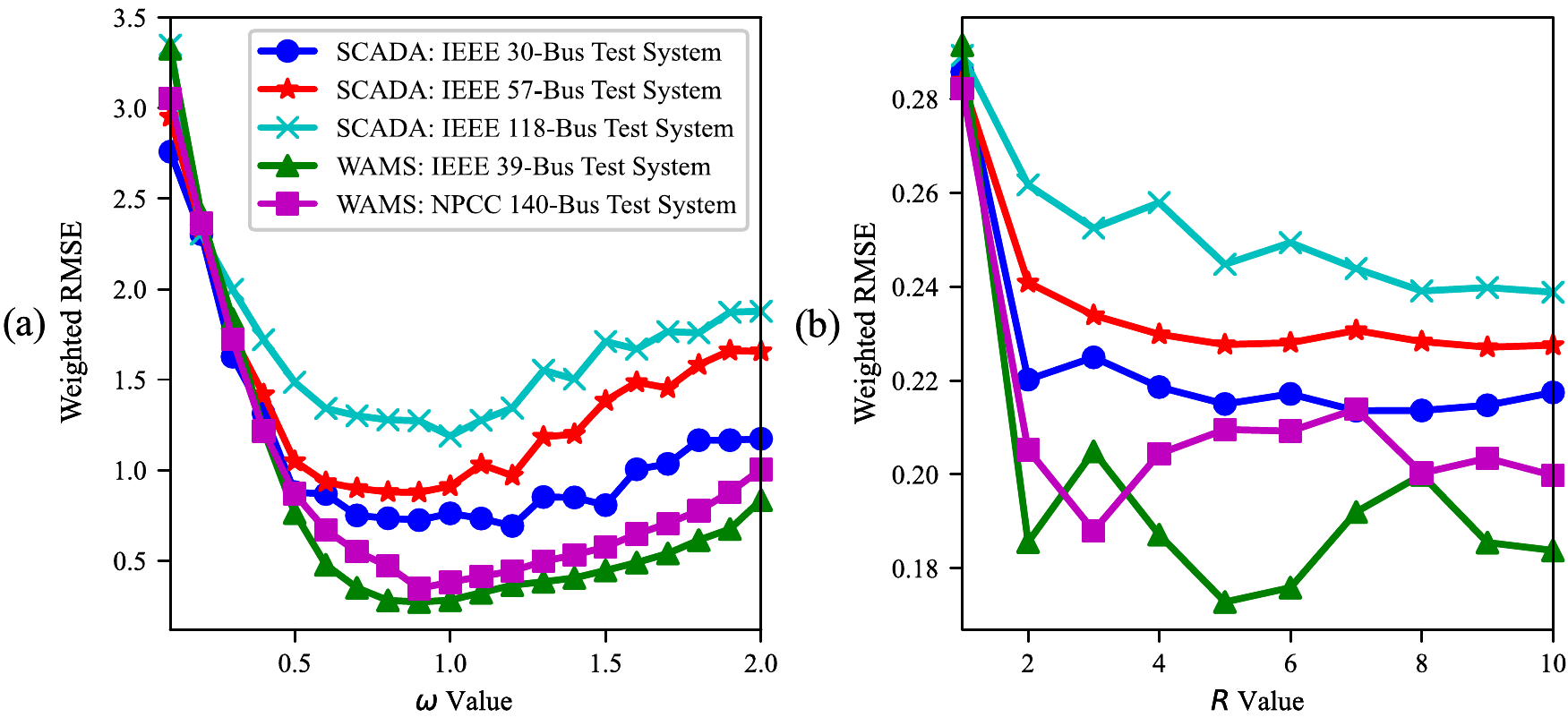}}
	\caption{The weighted RMSE under different (a) $\omega$ and (b) $R$.}
	\label{WR}
	\vspace{-0.1cm}
\end{figure}

It is necessary to select appropriate values of the hyperparameters $\omega$ in Stage 1 and $R$ in Stage 2. In this experiment, $\omega$ and $R$ are adjusted between $[\mathrm{0.1,2.0}]$ and $[\mathrm{1,10}]$, respectively, to determine the optimal values. The test dataset for $\omega$ contains outliers caused by step, ramp, and random attacks, while the dataset for $R$ contains RM and NM data loss. The average weighted RMSE is shown in Fig. \ref{WR}. It can be seen that the recovery error increases with the expansion of the system scale. Meanwhile, steady-state data recovery tends to have larger error than dynamic data recovery. In the guided conditional denoising diffusion process of Stage 1, as $\omega$ gradually increases, the diffusion generation trajectory of TSDM gradually shifts from random to deterministic, and the recovery error gradually decreases. Nonetheless, when $\omega$ is too large, the generation result is exactly the same as the original contaminated measurements $\boldsymbol{y}_0$, and the recovery error increases. In this way, it is appropriate to set $\omega$=1.0. In the diffusion-based imputation process of Stage 2, a larger $R$ will result in smaller errors but higher computational complexity. To this end, $R$ is set to 2.

\subsection{Robustness on FDIA and Data Losses}

\begin{figure}[!h]
	\vspace{-0.1cm}
	\centerline{\includegraphics[width=0.49\textwidth]{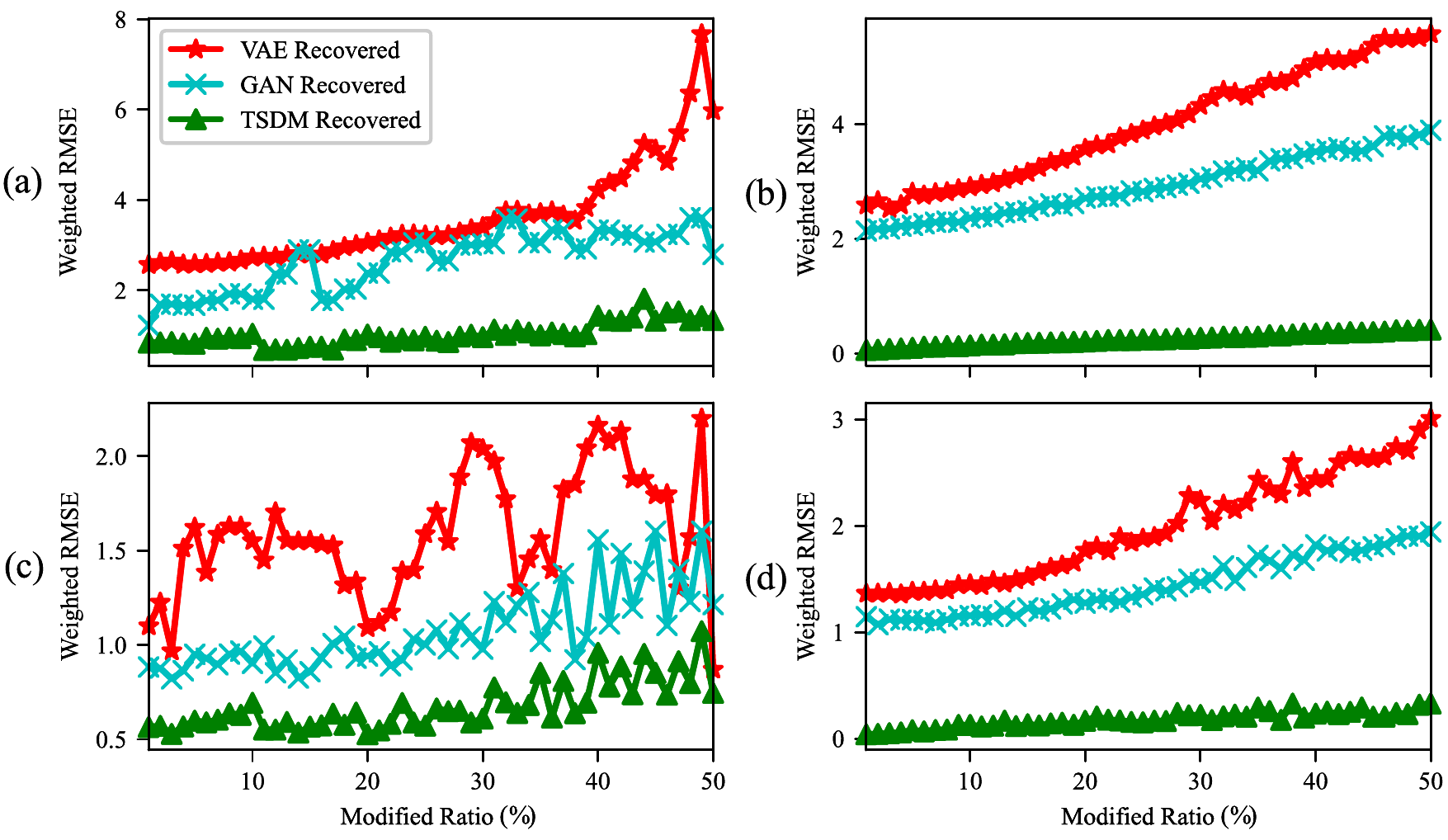}}
	\caption{SCADA FDIA (a) and data losses (b), WAMS FDIA (c) and data losses (d) recovery RMSE with different modified ratios.}
	\label{Ratio}
	\vspace{-0.1cm}
	\end{figure}

TSDM can also maintain better robustness in different scenarios. To verify this, the reconstruction errors of VAE, GAN, and TSDM algorithms are calculated at a modified ratio of 1\%-50\%, as illustrated in Fig. \ref{Ratio}. It can be seen that the data recovery error of various algorithms gradually increases as the proportion of anomalies increases. Nonetheless, the robustness performance of TSDM is better than the other two commonly used reconstruction algorithms. The proposed two-stage strategy effectively improves the robustness of TSDM.

\begin{table}[!h]\centering \footnotesize
	\vspace{-0.1cm}
	\caption{The recovery error and time consumption of DDPM, DDIM, and TSDM (weighted RMSE / time consumption (second))}
	\label{Timeliness}
	\renewcommand{\arraystretch}{1.0}
	\setlength\tabcolsep{0.5em}
	\begin{tabular}{cccccc}
	\toprule
	\toprule
     \multicolumn{1}{c}{Data type} & \multicolumn{1}{c}{\multirow{2}{*}{Model}} & \multicolumn{4}{c}{Subsequence Length}                                                 \\
     \cmidrule(lr){3-6}
     \multicolumn{1}{c}{$M \times T$} & \multicolumn{1}{c}{}                   & \multicolumn{1}{c}{$s$=10} & \multicolumn{1}{c}{$s$=20} & \multicolumn{1}{c}{$s$=50} &  \multicolumn{1}{c}{$s$=100}\\
	\midrule
	\multirow{3}{*}{\shortstack{SCADA:\\IEEE 30-bus\\$24 \times 96$}}                    & DDPM & -  & -    & -  & 4.35/9.22 \\
	& DDIM & 1.30/0.94   & 0.89/1.85  & 0.74/4.79 & 0.68/9.14 \\
	& TSDM & \textbf{0.69/0.91}  & \textbf{0.66/1.81}  & \textbf{0.65/4.49} & \textbf{0.64/9.19} \\
	\midrule
	\multirow{3}{*}{\shortstack{SCADA:\\IEEE 57-bus\\$48 \times 96$} }                    & DDPM &  -  & - & - & 4.30/17.61 \\
	& DDIM & 1.37/1.77  & 1.01/3.56 & 0.84/8.91 & 0.79/17.43 \\
	& TSDM & \textbf{0.82/1.76}  & \textbf{0.79/3.47}  & \textbf{0.78/8.80} & \textbf{0.76/17.59}  \\
	\midrule
	\multirow{3}{*}{\shortstack{SCADA:\\IEEE 118-bus\\$96 \times 96$} }                    & DDPM &  -  & - & - & 5.29/36.07 \\
	& DDIM & 1.35/3.53  & 1.00/7.13 & 0.86/18.09 & 0.83/35.35 \\
	& TSDM & \textbf{0.81/3.63}  & \textbf{0.79/7.05}  & \textbf{0.77/17.90} & \textbf{0.76/35.99}  \\
	\midrule
	\multirow{3}{*}{\shortstack{WAMS:\\IEEE 39-bus\\$48 \times 120$} }                    & DDPM &  -  & - & - & 4.75/21.43 \\
	& DDIM & 1.18/2.18  & 0.79/4.35 & 0.51/10.87 & 0.41/21.62 \\
	& TSDM & \textbf{0.43/2.23}  & \textbf{0.39/4.44}  & \textbf{0.36/10.91} & \textbf{0.34/21.75}  \\
	\midrule
	\multirow{3}{*}{\shortstack{WAMS:\\NPCC 140-bus\\$120 \times 120$}}                    & DDPM &  -  & - & - & 3.87/51.02 \\
	& DDIM & 1.20/5.16  & 0.81/10.44 & 0.57/25.79 & 0.51/51.31 \\
	& TSDM & \textbf{0.53/5.21}  & \textbf{0.48/10.59}  & \textbf{0.45/26.09} & \textbf{0.44/51.36}  \\			
	\bottomrule
	\bottomrule
	\end{tabular}
	\vspace{-0.1cm}
	\end{table}

\subsection{Timeliness with Acceleration Strategy}

DDIM with precise estimated mean and optimal variance is one of the contributions of TSDM, which can effectively mitigate one of the drawbacks of diffusion models, i.e., more prolonged time consumption than some commonly used AI models. The time consumption and weighted RMSE of DDPM with original timestep length, DDIM with different subsequence lengths, and TSDM algorithm with optimal variance in different test systems with attack region measurement dimensionality ($M \times T$) are shown in Table \ref{Timeliness}. The diffusion generation process of the original DDPM is completely random and cannot be directly applied to measurement rectification or state reconstruction, resulting in significant recovery errors. However, under the same subsequence length, the reconstruction RMSE of TSDM is much lower than the original conditional DDIM. Meanwhile, under the same reconstruction precision, the time consumption of TSDM is much less (about 10\%) than that of the original diffusion models. Moreover, the proposed TSDM can flexibly partition attack areas of a power system and reduce data dimensionality to achieve faster data recovery.

%section Case Studies (end)

\section{Conclusion} % (fold)
\label{sec:Conclusion}
This paper proposes an improved two-stage power system measurement recovery model (TSDM) based on denoising diffusion models to mitigate the influence of cyber-physical uncertainties. The proposed detection and imputation based TSDM can effectively extract the spatio-temporal correlations and measurement coordinations between data points and use the extracted patterns to guide the process of outlier elimination and missing data supplementation. Compared with existing reconstruction methods, the proposed model demonstrates high recovery accuracy and strong robustness under highly random and nonlinear dynamics of power systems. This is evidenced through extensive case studies based on SCADA and WAMS data. Nonetheless, the exploration of real-time data recovery leveraging denoising latent diffusion models \cite{rombach2022high} warrants further investigation.

% section conclusion (end)

\appendix

\noindent \textit{A. System Uncertainty}

The matrices $\bm{A}_{Gi}$, $\bm{B}_{Gi}$, and $\bm{C}_{Gi}$ are denoted as	
\begin{equation} 
	\begin{aligned}
		\bm{A}_{Gi}= \begin{bmatrix}
			0 & 1 & 0 & 0\\ 
			0 & -\frac{D_{Gi}}{2H_{Gi}} & 0 & 0\\ 
			0 & 0 & -\frac{1}{T'_{qGi0}} & 0\\ 
			 0& 0 & 0 & -\frac{1}{T'_{dGi0}}
			\end{bmatrix},
	\end{aligned}
\end{equation}	
\begin{equation} 
	\begin{aligned}
		& \bm{B}_{Gi}= \begin{bmatrix}
			0 & 0\\ 
			-\frac{1}{2H_{Gi}} & 0\\ 
			0 & 0\\ 
			0 & 0
			\end{bmatrix} \\
			& + \begin{bmatrix}
			0 &0 \\ 
			0 & 0\\ 
			-\frac{X_{qGi}-X'_{qGi}}{T'_{qGi0}} &0 \\ 
			 0& -\frac{X_{dGi}-X'_{dGi}}{T'_{dGi0}}
			\end{bmatrix}\times \begin{bmatrix}
			U_{qGi} & U_{dGi}\\ 
			 -U_{dGi}& U_{qGi}
			\end{bmatrix}^{-1}, 
	\end{aligned}
\end{equation}
\begin{equation} 
	\begin{aligned}
		\bm{C}_{Gi}= \begin{bmatrix}
			0 & 0\\ 
			\frac{1}{2H_{Gi}} & 0 \\ 
			0 & 0 \\ 
			0 & \frac{1}{T'_{dGi0}}
			\end{bmatrix}.
	\end{aligned}
\end{equation}	

By combining generator dynamics Eq. \eqref{dgdr} and network constraints Eq. \eqref{ncr}, the power system state fluctuation can be derived as 
\begin{equation} 
	\begin{aligned}
		\begin{bmatrix}
			\Delta \bm{s}_{G,k}\\ 
			\Delta \bm{\theta_k}\\ 
			\Delta \bm{U}_k
			\end{bmatrix} = & \underbrace{\begin{bmatrix}
				\Delta t \bm{A}_G +1 & \bm{0} &\bm{0} \\ 
				\bm{0} & \bm{0} & \bm{0}\\ 
				\bm{0} & \bm{0} & \bm{0}
			   \end{bmatrix}}_{\bm{A}} \begin{bmatrix}
			   \Delta \bm{s}_{G,k-1}\\ 
			   \Delta \bm{\theta}_{k-1}\\ 
			   \Delta \bm{U}_{k-1}
			   \end{bmatrix}\\
			&+\underbrace{\begin{bmatrix}
			(\Delta t \bm{B}_G)^*\\ 
			\bm{D}^{-1}
			\end{bmatrix}}_{\bm{B}}\begin{bmatrix}
			\Delta  \bm{P}_{G,k}\\ 
			\Delta  \bm{Q}_{G,k}
			\end{bmatrix}\\
			&+ \underbrace{\begin{bmatrix}
			\Delta t \bm{C}_G\\ 
			\bm{0}
			\end{bmatrix}}_{\bm{C}}\Delta \bm{u}_k - \underbrace{\begin{bmatrix}
			\bm{0}\\ 
			\bm{D}^{-1}
			\end{bmatrix}}_{\bm{D}}\begin{bmatrix}
			\Delta  \bm{P}_{L,k}\\ 
			\Delta  \bm{Q}_{L,k}
			\end{bmatrix},
	\end{aligned} 
\end{equation}
where $(\Delta t \bm{B}_G)^*$ means that $\Delta t \bm{B}_G$ is resheduled by $(.)^*$.

\noindent \textit{B. Optimal Variance}

Eq. \eqref{DDIMreverse} can be transformed into a sampling form 
\begin{equation}
	\begin{aligned}
		\boldsymbol{x}_{n-1} & \approx \frac{\sqrt{1-\alpha_{n-1}-\sigma_n^2  }}{\sqrt{1-\alpha_n}}\boldsymbol{x}_n+\gamma_n(\bar{\mu}(\boldsymbol{x}_n)+\bar{\sigma}_n\epsilon_2)+\sigma_n\epsilon_1 \\
		& =\frac{\sqrt{1-\alpha_{n-1}-\sigma_n^2  }}{\sqrt{1-\alpha_n}}\boldsymbol{x}_n+\gamma_n\bar{\mu}(\boldsymbol{x}_n)+(\underbrace{ \sigma_n\epsilon_1+\gamma_n\bar{\sigma}_n\epsilon_2}_{\sim \sqrt{\sigma_n^2+\gamma_n^2\bar{\sigma}^2_n}\epsilon }),
	\end{aligned}
\end{equation}
where $\epsilon, \epsilon_1, \epsilon_2$ are i.i.d. as $\mathcal{N}(\bm{0},\boldsymbol{I})$. Evidently, $p(\boldsymbol{x}_{n-1}|\boldsymbol{x}_n)$ is closer to the Normal distribution with mean $\frac{\sqrt{1-\alpha_{n-1}-\sigma_n^2  }}{\sqrt{1-\alpha_n}}\boldsymbol{x}_n+\gamma_n\bar{\mu}(\boldsymbol{x}_n)$ and covariance $(\sigma_n^2+\gamma_n^2\bar{\sigma}^2_n ) \boldsymbol{I}$, and even if $\sigma_n=0$, the estimated variance of the reverse process is not zero. In summary, it is significant to estimate the optimal variance $(\gamma_n^2 \bar{\sigma}_n^2) \boldsymbol{I}$. According to Eq. \eqref{meanofx0}, the covariance of $\boldsymbol{x}_n$ can be represented as 
\begin{equation} \label{covariance}
	\begin{aligned}
		\Sigma(\boldsymbol{x}_n) &=\mathbb{E}_{\boldsymbol{x}_0\sim p(\boldsymbol{x}_0|\boldsymbol{x}_n)}\left [ (\boldsymbol{x}_0-\bar{\mu}(\boldsymbol{x}_n)) (\boldsymbol{x}_0-\bar{\mu}(\boldsymbol{x}_n))^\top \right ] \\
		&= \frac{1}{\alpha_n }\underbrace{\mathbb{E}_{\boldsymbol{x}_0\sim p(\boldsymbol{x}_0|\boldsymbol{x}_n)}\left [ (\boldsymbol{x}_0 -\sqrt{\alpha_n }\boldsymbol{x}_0)(\boldsymbol{x}_0 -\sqrt{\alpha_n }\boldsymbol{x}_0)^\top \right ]}_{\boldsymbol{e}_n} \\
		&-\frac{1-\alpha_n}{\alpha_n} \epsilon_{\theta} (\boldsymbol{x}_n,n)\epsilon_{\theta} (\boldsymbol{x}_n,n)^\top.	
	\end{aligned}
\end{equation}
According to the derivation of DDIM, it can be inferred that $p(\boldsymbol{x}_n|\boldsymbol{x}_0)=\mathcal{N}(\boldsymbol{x}_n; \sqrt{\alpha_n}\boldsymbol{x}_0,(1-\alpha_n)\boldsymbol{I})$ and the calculation result of $\boldsymbol{e}_n$ can be represented as $(1-\alpha_n) \boldsymbol{I}$. Collectively, Eq. \eqref{covariance} can be rewritten as 
\begin{equation} \label{covariancedeep}
	\begin{aligned}
		\Sigma_n &=\mathbb{E}_{\boldsymbol{x}_n \sim p(\boldsymbol{x}_n)}\left [ \Sigma (\boldsymbol{x}_n) \right ] \\
		&=\frac{1-\alpha_n}{\alpha_n}  \left \{ \boldsymbol{I}-\mathbb{E}_{\boldsymbol{x}_n\sim p(\boldsymbol{x}_n)} \left [ \epsilon_{\theta}(\boldsymbol{x}_n,n)\epsilon_{\theta}(\boldsymbol{x}_n,n)^\top \right ] \right \}  .	
	\end{aligned}
\end{equation}
Ultimately, both sides of Eq. \eqref{covariancedeep} are traced and divided by the dimensionality $d$ $(M \times T)$ of power system measurements, then the optimal variance can be expressed as 
\begin{equation}
	\begin{aligned}
		\bar{\sigma}_n^2 = \frac{1-\alpha_n }{\alpha_n} \left \{  1- \frac{1}{d} \mathbb{E}_{\boldsymbol{x}_n\sim p(\boldsymbol{x}_n)}\left [ \left \| \epsilon_{\theta}(\boldsymbol{x}_n,n) \right \|^2_2  \right ] \right \}.	
	\end{aligned}
\end{equation}

\bibliographystyle{IEEEtran}
\bibliography{IEEEabrv,bibi}

\begin{IEEEbiography}[{\includegraphics[width=1in,height=1.25in,clip,keepaspectratio]{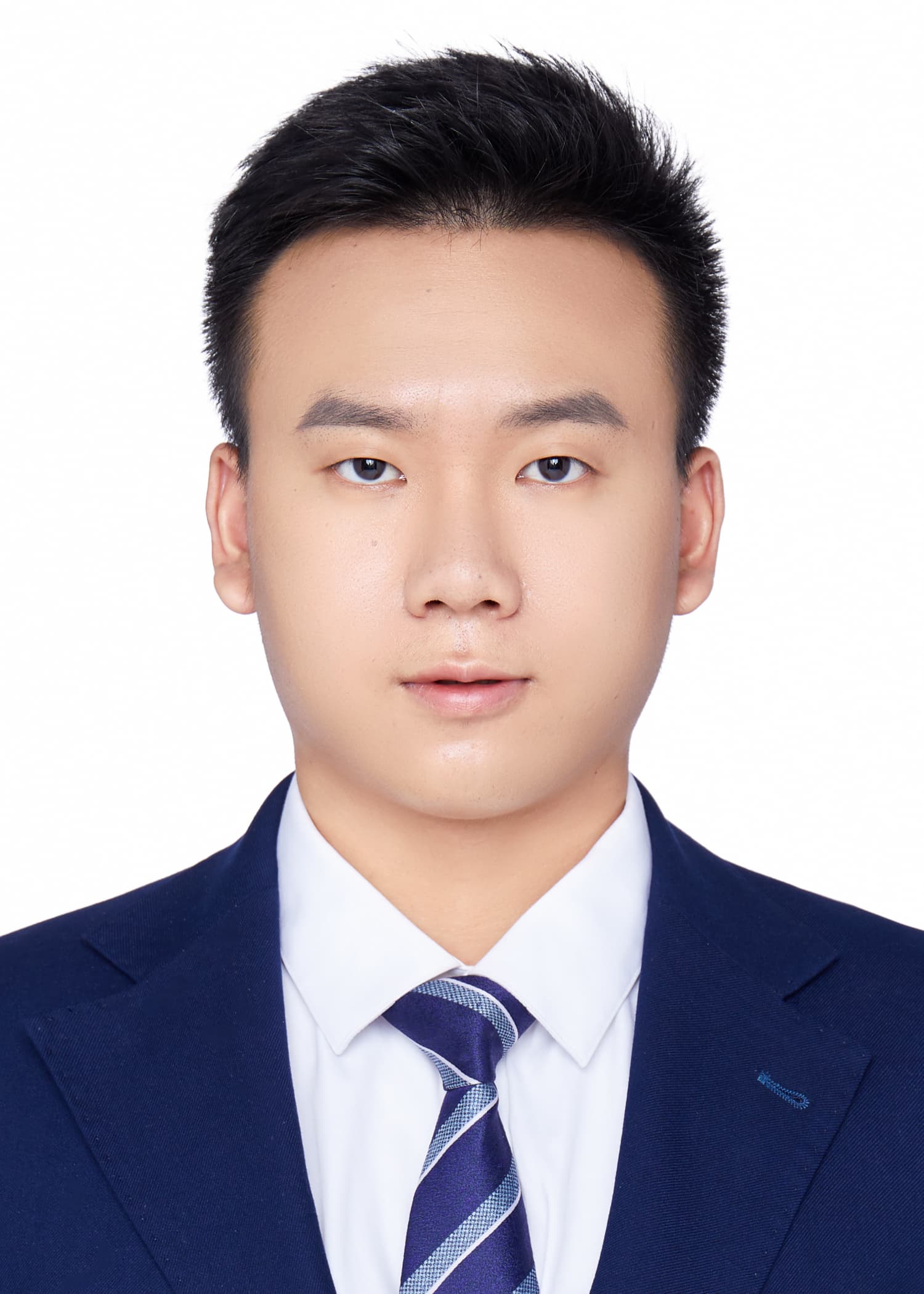}}]{Jianhua Pei} (Student Member, IEEE) received the B.Eng. degree in electrical engineering from Huazhong University of Science and Technology (HUST), Wuhan, China, in 2019. He is currently pursuing his Ph.D. degree in electrical engineering at HUST. He is also a visiting Ph.D. student with the Department of Electrical Engineering and Computer Science, Lassonde School of Engineering, York University, Canada, in 2024.

His research interests include power system data quality improvement, power system dynamics, power system cybersecurity, and artificial intelligence applications for communications.
		
\end{IEEEbiography}
		
\begin{IEEEbiography}[{\includegraphics[width=1in,height=1.25in,clip,keepaspectratio]{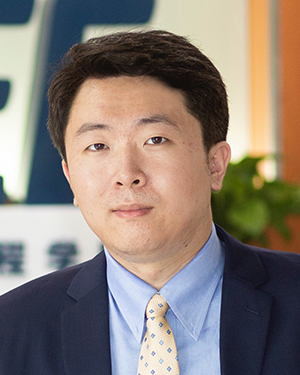}}]{Jingyu Wang} (Member, IEEE) received the B.Eng. degree and the Ph.D. degree in electrical engineering from Huazhong University of Science and Technology (HUST), Wuhan, China, in 2015 and 2021, respectively. He is currently a postdoctoral research fellow with the School of Electrical and Electronic Engineering and the School of Cyber Science and Engineering, HUST. From 2019 to 2020, he was a Visiting Scholar with Virginia Polytechnic Institute and State University, Blacksburg, USA.
	
His research interests include power system cybersecurity, artificial intelligence applications, and novel technologies in power system monitoring and control.
\end{IEEEbiography}

\begin{IEEEbiography}[{\includegraphics[width=1in,height=1.25in,clip,keepaspectratio]{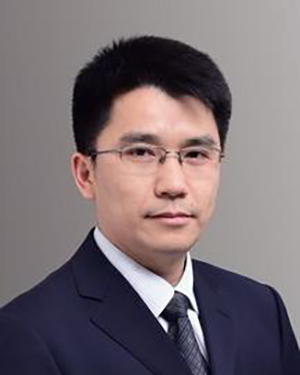}}]{Dongyuan Shi} (Senior Member, IEEE) received the B.S. and Ph.D. degrees in electrical engineering from Huazhong University of Science and Technology (HUST), China, in 1996 and 2002, respectively. From 2007 to 2009, he was a Visiting Scholar with Cornell University, Ithaca, NY. He is currently a professor with the School of Electrical and Electronic Engineering, HUST.
	
His research interests include power system analysis and computation, cybersecurity, and software technology.
\end{IEEEbiography}

\begin{IEEEbiography}[{\includegraphics[width=1in,height=1.25in,clip,keepaspectratio]{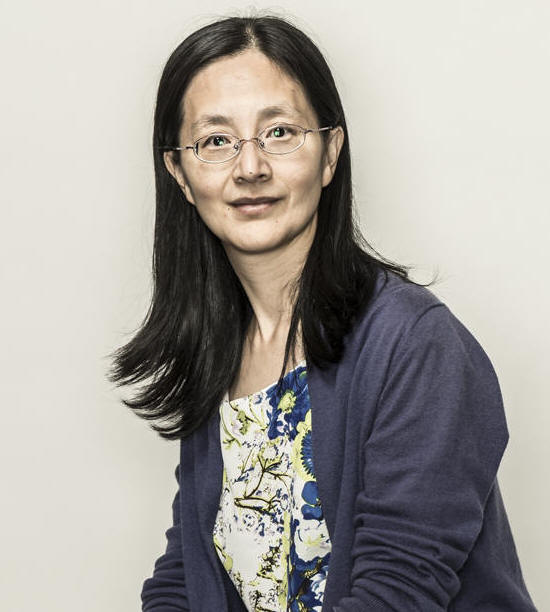}}]{Ping Wang} (Fellow, IEEE) is a Professor at the Department of Electrical Engineering and Computer Science, York University, and a Tier 2 York Research Chair. Prior to that, she was with Nanyang Technological University, Singapore, from 2008 to 2018. 

Her research interests are mainly in radio resource allocation, network design, performance analysis and optimization for wireless communication networks, mobile cloud computing and the Internet of Things. Her recent works focus on integrating Artificial Intelligence (AI) techniques into communications networks. Her scholarly works have been widely disseminated through top-ranked IEEE journals/conferences, received 29,000+ citations, and received the Best Paper Awards from IEEE prestigious conference WCNC in 2012, 2020 and 2022, from IEEE Communication Society: Green Communications \& Computing Technical Committee in 2018, from IEEE flagship conference ICC in 2007. She has been serving as an associate editor-in-chief for IEEE Communications Surveys \& Tutorials and an editor for several reputed journals including IEEE Transactions on Wireless Communications. She is an IEEE Fellow and a Distinguished Lecturer of the IEEE Vehicular Technology Society.
\end{IEEEbiography}

\end{document}